\documentclass[11pt]{article}
\usepackage{geometry}
\usepackage[dvipsnames]{xcolor}
\usepackage{array}
\usepackage{bbding}
\usepackage{multirow}
\usepackage[fleqn]{amsmath}
\usepackage{amssymb}
\usepackage{amsthm}
\usepackage{graphicx}
\usepackage[authoryear]{natbib}
\usepackage{lscape}
\usepackage[hyphens]{url}
\usepackage[hidelinks]{hyperref}
\usepackage{comment}
\usepackage{bm}
\usepackage[title]{appendix}
\usepackage{longtable}
\usepackage{mathtools}
\usepackage{chngcntr}
\usepackage{authblk}
\usepackage{babel}
\usepackage{dsfont}
\usepackage{subcaption}
\usepackage{booktabs}
\usepackage{makecell}
\usepackage{array}
\usepackage{fancyhdr}
\usepackage{cleveref}
\usepackage{abstract}
\usepackage{threeparttable}
\setcitestyle{maxnames=5}

\makeatletter
\allowdisplaybreaks
\date{}

\fancypagestyle{plain}{%
	\fancyhf{} 
	
	\lhead{\footnotesize \textcolor{gray}{\textit{\runningtitle -- paper under review}}}
	\rhead{\footnotesize \textcolor{gray}{\thepage}}
}
\makeatletter
\def\blfootnote{\xdef\@thefnmark{}\@footnotetext}
\makeatother

\makeatletter
\def\titlepageext{
	\begin{center}
	{\parindent0pt
		\rule{0.9\textwidth}{1pt}
		\begin{minipage}[t]{0.25\textwidth}
			\small {\it Keywords:}\\
			\keywords
		\end{minipage}%
		\hspace{3mm}
		\begin{minipage}[t]{0.6\textwidth}
			\small \abstract
		\end{minipage}%
\vspace{2mm}
		\rule{0.9\textwidth}{1pt}
	}
	\end{center}

	\blfootnote{* Corresponding author. E-mail address: \href{mailto:\corresemail}{\corresemail}.}
}
\makeatother

\newtheorem{prop}{Proposition}

\DeclareMathOperator*{\E}{\mathcal{E}}
\DeclareMathOperator*{\G}{\mathcal{G}}
\DeclareMathOperator*{\N}{\mathcal{N}}

\DeclareMathOperator*{\V}{\mathcal{V}}

\title{Graph neural networks for residential location choice: connection to classical logit models}

\def\runningtitle{GNNs for residential location choice}

\author[a]{Zhanhong Cheng}
\author[b]{Lingqian Hu}
\author[c]{Yuheng Bu}
\author[a]{Yuqi Zhou}
\author[a*]{Shenhao Wang}
\affil[a]{Department of Urban and Regional Planning, University of Florida, Gainesville, FL, USA}
\affil[b]{Department of Landscape Architecture and Urban Planning, Texas A\&M University, Texas, USA}
\affil[c]{Department of Computer Science, University of California, Santa Barbara, CA, USA}
\def\corresemail{shenhaowang@ufl.edu}

\def\abstract{Researchers have adopted deep learning for classical discrete choice analysis as it can capture complex feature relationships and achieve higher predictive performance. However, the existing deep learning approaches cannot explicitly capture the relationship among choice alternatives, which has been a long-lasting focus in classical discrete choice models. To address the gap, this paper introduces Graph Neural Network (GNN) as a novel framework to analyze residential location choice. The GNN-based discrete choice models (GNN-DCMs) offer a structured approach for neural networks to capture dependence among spatial alternatives, while maintaining clear connections to classical random utility theory. Theoretically, we demonstrate that the GNN-DCMs incorporate the nested logit (NL) model and the spatially correlated logit (SCL) model as two specific cases, yielding novel algorithmic interpretation through message passing among alternatives' utilities. Empirically, the GNN-DCMs outperform benchmark MNL, SCL, and feedforward neural networks in predicting residential location choices among Chicago's 77 community areas. Regarding model interpretation, the GNN-DCMs can capture individual heterogeneity and exhibit spatially-aware substitution patterns. Overall, these results highlight the potential of GNN-DCMs as a unified and expressive framework for synergizing discrete choice modeling and deep learning in the complex spatial choice contexts.}

\def\keywords{Residential location choice \\ Spatial choice model\\ Graph neural network \\ Discrete choice model \\ Nested logit model \\}

\date{}

\begin{document}

\maketitle
\titlepageext

\section{Introduction}
It is essential to understand individual households' residential location choices for transportation planning \citep{ben1998integration}, urban development \citep{bhat2007comprehensive}, and economic policy \citep{bayoh2006determinants}. Residential location choices are typically framed as discrete choice modeling (DCM) tasks. But unlike other widely-studied DCMs, such as travel mode choice, residential location choice is more challenging because it involves a \textit{large} set of \textit{spatially correlated} alternatives (e.g., a neighborhood). Such complexity introduces several challenges for classical DCMs from the 1970s through the 2010s. First, the spatial correlation among alternatives violates the Independence of Irrelevant Alternatives (IIA) assumption inherent in the Multinomial Logit (MNL) model. Second, the large size of the alternative set makes it difficult to specify correlation structures. Third, residential choices often involve complex interactions between individual characteristics and location attributes. In the classical framework, researchers have been seeking to address these challenges using the nested logit (NL) model \citep{mcfadden1978modelling}, spatially correlated logit (SCL) model \citep{bhat2004mixed}, and its extensions \citep{sener2011accommodating, perez2022spatially}.

In recent years, neural networks (NN) have gained increasing attention because they can capture non-linear relationships and complex interactions between individual and alternative attributes, outperforming traditional DCMs in predictive accuracy \citep{wang2024comparing}. However, the majority of the existing studies introduce only artificial neural networks (ANNs) and their variants, thus failing to explicitly capture the \textit{dependencies among choice alternatives}. The existing ANN-based discrete choice models assume independent alternatives \citep{wang2020deepASU}, implicitly encode alternatives' correlations by feeding the attributes of all alternatives \citep{wang2020deep, wong2021reslogit}, or embed ANNs within traditional DCMs to facilitate utility specification \citep{sifringer2020enhancing}. None of them can explicitly capture alternatives' dependencies while maintaining a clear connection to classical discrete choice theory.

\begin{figure}[htbp]
    \centering
    \begin{subfigure}{\textwidth}
        \centering
        \includegraphics[scale=0.8]{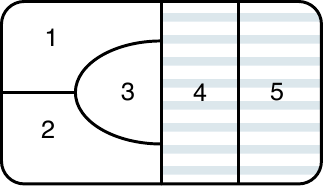}
        \caption{Spatial configuration of residential locations.}
        \label{fig:toy}
    \end{subfigure}

    \begin{subfigure}{\textwidth}
        \centering
        \includegraphics[scale=0.8]{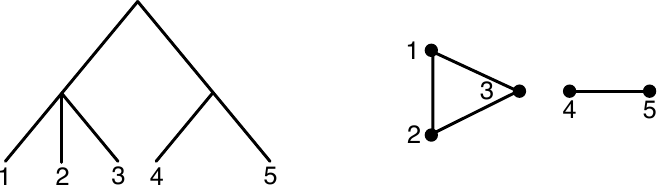}
        \caption{The nested structure of the NL model (left) and its corresponding alternative graph (right) for the GNN.}
        \label{fig:toy_nll}
    \end{subfigure}

    \begin{subfigure}{\textwidth}
        \centering
        \includegraphics[scale=0.8]{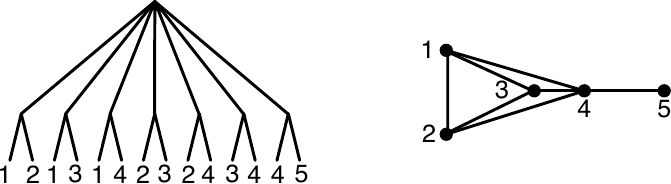}
        \caption{The nested structure of the SCL model (left) and its corresponding alternative graph (right) for the GNN.}
        \label{fig:toy_scl}
    \end{subfigure}

    \caption{Illustration of the alternative graphs in the NL and SCL models for residential location choice.}
    \label{fig:intro}
\end{figure}

To address such limitations, this paper proposes a Graph Neural Network framework for discrete choice modeling (GNN-DCMs), specifically designed for large sets of spatially correlated alternatives such as residential location choice. Instead of embedding neural networks directly in a Generalized Extreme Value (GEV) model \citep{sifringer2020enhancing}, the GNN-DCMs can capture alternatives' correlation through message passing between utilities of alternatives. Figure.~\ref{fig:intro} illustrates a novel concept, i.e., alternative graph, which provides a new interpretation for NL and SCL models as message passing among alternative utilities. Combining the alternative graph and GNNs, we demonstrate theoretically that GNN-DCMs incorporate classical NL, SCL, and ASU-DNN \citep{wang2020deepASU} as three special cases. Additionally, we derive the relationship between cross-elasticities and the relative positions of two alternatives on the graph, revealing the spatial-aware substitution patterns of the GNN-DCMs. Using the sample of 3,838 households' residential location choices among the 77 community areas in Chicago, our GNN-DCMs outperform MNL, SCL, and ANN benchmarks on predictive performance. This case study also demonstrates the model's ability to capture individual heterogeneity and spatial-aware substitution patterns.

This study seeks to advance the methodological frontier of discrete choice modeling by introducing a GNN framework to capture complex spatial correlation among alternatives, such as housing location choices. The key contributions are as follows:
\begin{itemize}
    \item Proposing a GNN-DCM framework that handles spatially correlated alternatives through message passing, scales efficiently with large alternative sets, and captures complex non-linear interactions in individual and alternative attributes.
    \item Establishing theoretical connections between GNN-DCMs and classical DCMs, illustrating that the GNN-DCM framework generalizes both traditional DCM and existing ANN models.
    \item Demonstrating GNN-DCMs' strong predictive performance, spatially-aware substitution patterns, and the ability to reveal non-linearity and individual heterogeneity in choice behavior.
\end{itemize}

The remainder of this paper is organized as follows. Section~\ref{sec:review} reviews relevant literature on spatially correlated discrete choice models, including both classical and neural network-based approaches. Section~\ref{sec:methodology} presents the GNN-DCM framework, detailing its structure and theoretical properties. Section~\ref{sec:case study} applies the framework to a residential location choice problem in Chicago and compares its performance to baseline models. Finally, Section~\ref{sec:conclusions} concludes the paper and outlines future research directions.

\section{Literature Review}\label{sec:review}
\subsection{Classical discrete choice models}\label{sec:review_classical}
The Multinomial Logit (MNL) model \citep{mcfadden1973conditional}  has been widely applied in empirical studies to examine the factors influencing residential location choice \citep{weisbrod1980tradeoffs,guevara2006endogeneity,hu2019housing}. Typically, the MNL model assumes that the random utility components are independent and identically distributed (IID) across alternatives, thus ignoring potential spatial correlations in the choice set. To address this limitation, a range of models have been proposed for modeling location choice, including the Multinomial Probit (MNP) model \citep{bolduc1991multinomial}, mixed Logit models \citep{bhat2004mixed}, and Generalized Extreme Value (GEV) models \citep{mcfadden1978modelling}. Probit and mixed logit models can capture spatial correlations with the correlated error structure or utility components. However, they often require computationally intensive numerical integration methods for estimation, which can be challenging with large alternative sets. In contrast, GEV models are particularly attractive because many of them offer closed-form choice probabilities and scale well with a large set of correlated alternatives.

GEV represents a broad family of models, which capture the dependency of choice alternatives through a joint generalized extreme value distribution over the unobserved utility components \citep{train2009discrete}. The NL model \citep{ben1973structure,mcfadden1978modelling,lee2010residential} is the very first member in the GEV family, and has been used for residential location choice (illustrated on the left side of Figure.~\ref{fig:toy_nll}). Building upon the generalized nested model proposed by \cite{wen2001generalized}, \citet{bhat2004mixed} introduced the Spatially Correlated Logit (SCL) model for residential location choice, where two alternatives are grouped into the same nest if they are spatially adjacent (as shown in Figure.~\ref{fig:toy_scl}, left panel). The authors also developed a mixed SCL model to capture individual-level heterogeneity. Further extensions include \citet{sener2011accommodating}, who generalized the SCL model by parameterizing the nest allocation parameters using an MNL structure, allowing for more flexible and data-driven correlation patterns. Recently, \cite{perez2022spatially} combined the nested logit model with the SCL model, capturing the spatial correlations for alternatives in the same nest. In short, the GEV model family, such as NL and SCL models, are computationally efficient approaches for capturing spatial dependency among choice alternatives.

Our study focuses on using deep learning to capture the spatial dependence across the discrete alternatives, as tackled by the classical NL and SCL models \citep{mcfadden1978modelling, bhat2004mixed, sener2011accommodating}. In this case, spatial dependence manifests as correlation across alternatives, rather than across individuals or independent variables. This dependence of choice alternatives poses a challenge categorically different from that of individuals. For example, prior studies on dependencies among individuals analyze transportation mode choice assuming that individuals residing nearby prefer similar transportation modes \citep{bhat2000multi, dugundji2005discrete, goetzke2008network} or land use categories  \citep{bina2006location, wang2012dynamic}. In such cases, spatial dependence may arise from spatial spillover effects—where spatial factors (e.g., multiple transit stops) influence one another \citep{lesage2009introduction}—or from self-selection effects, where individuals with similar preferences (e.g., active travel) or social connections tend to cluster geographically \citep{bhat2007comprehensive, bhat2009copula}. Nonetheless, our study does not seek to capture the dependence among individuals, as it has been addressed broadly by social network literature.

\subsection{Neural network approaches}\label{sec:review_nn}
Researchers have adopted artificial neural networks (ANNs) to tackle discrete choice modeling tasks, outperforming the classical models by capturing complex relationship between individuals and alternatives. In these ANNs, the input layer consists of attributes of both individuals and alternatives, while the output layer produces utility values. These utilities are then transformed into choice probabilities using the softmax function, the same as the MNL formulation. Because of their ability to model complex non-linear relationships between features, ANNs often outperform traditional discrete choice models in predictive accuracy \citep{wang2024comparing}. ANNs have been applied to discrete choice problems since the 1990s \citep{kumar1995empirical, agrawal1996market}. Although early applications were relatively straightforward applications of ANNs, recent research interests are increasingly growing in terms of synergizing ANNs and classical DCMs. For example, the efforts include extracting economic information from ANNs \citep{wang2020deep}, enhancing ANNs' interpretability \citep{feng2024deep, wang2020deepASU}, or using ANNs to capture individual heterogeneity.

Despite these advances, it remains underexplored how to capture the correlations among alternatives within the deep learning framework. For instance, \cite{wang2020deepASU} employs alternative-specific utility functions, which only preserve the independence among alternatives without addressing the alternative dependence. \cite{wong2021reslogit} proposed ResLogit, where the observed utility is defined as the sum of an alternative-specific MNL component and a cross-alternative ANN component, while the alternative structure is not studied. \cite{sifringer2020enhancing} used ANNs as a part of the utility in a NL model--using the classical GEV model as a backbone to model alternative correlations. These methods for capturing alternative correlations primarily rely on feeding the attributes of all alternatives into the network. Such approaches do not leverage prior knowledge of the alternative correlation, thus exhibiting low inductive bias and increasing the risk of overfitting, especially in large choice sets.

It is possible to analyze the alternative correlation through a network perspective and GNN models, although such a possibility has rarely been explored yet \citep{hamilton2017inductive, kipf2016semi, velivckovic2018graph}.
GNNs can capture network dependence through the graph-based message passing algorithm. They have demonstrated strong performance in a variety of domains and begin to be applied to DCMs. \cite{tomlinson2024graph} proposed a model with individual-specific coefficients, as an analogy to mixed logit models in the utility function, with a graph regularizer encouraging similarity among socially connected individuals. The most relevant work to our approach is by \cite{villarraga2025designing}, who used Graph Convolutional Networks (GCNs) \citep{kipf2016semi} to capture social network effects. Their model seeks to incorporate individual correlations, rather than the alternative correlation as in our study. To the best of our knowledge, our work is the first to apply GNNs to the DCM tasks by structurally analyzing a large number of spatially correlated alternatives, while retaining the economic interpretation by analyzing the substitution patterns and elasticities.

\section{Methodology: GNN-DCM Framework}
\label{sec:methodology}

This section proposes a GNN-DCM framework for residential location choice with spatial correlations. To capture dependence among alternatives, we introduce the concept of an alternative graph, which enables the integration of GNNs into discrete choice modeling through a network approach. We demonstrate that this GNN-DCM framework contains certain GEV and ANN models as specific cases. Finally, we derive the elasticities of GNN-DCMs, which help interpret the model's predictions and provide insights into spatially-aware substitution patterns.

\subsection{Alternative graph}\label{sec:alternative_graph}
Alternative graph enables us to capture the alternative dependence through GNNs. Specifically, this alternative graph is defined as $\G = (\V, \E)$, where $\V$ represents the set of choice alternatives, (e.g., residential location), and $\E$ denotes the set of edges that define the relationships among these alternatives. If two alternatives are independent, no edge exists between them; otherwise, they are connected by an edge. The edge set $\E$ is represented by an adjacency matrix $\mathbf{A}_{\left|\V\right| \times \left|\V\right|}$, where $\mathbf{A}[i,j] = 1$ if $(i,j)\in \E$ and $\mathbf{A}[i,j] = 0$ otherwise.
For residential location choice, edges can be defined based on geographical adjacency. For example, residential location alternatives in Figure.~\ref{fig:toy} can be represented as an alternative graph with a structure shown on the right side of Figure.~\ref{fig:toy_scl}. Currently, the alternative graph incorporates only undirected edges to represent these connections, which ensures a symmetric adjacency matrix.

Let $\mathbf{x}_{ni}$ denote the vector of explanatory variables of alternative $i$ and individual $n$. In the context of residential location choice, $\mathbf{x}_{ni}$ encompasses a set of individual- and alternative-specific attributes, such as household income, housing value, and distance to work. Consequently, when an individual $n$ makes the choice, each node in the alternative graph is associated with an attribute vector specific to that individual.

\subsection{Utility specification using GNN and alternative graphs}\label{sec:gnn}
Recall the random utility maximization (RUM) framework, where the utility of alternative $i$ for individual $n$ contains a deterministic component $V_{ni}$ and a random component $\varepsilon_{ni}$, i.e., $U_{ni} = V_{ni} + \varepsilon_{ni}$. Under the assumption that individuals choose the alternative with the highest utility, the probability that individual $n$ selects alternative $i$ is given by
\begin{equation*}
    P_{ni} = P(U_{ni}>U_{nj},\forall j\neq i).
\end{equation*}

When assuming the random components $\varepsilon_{ni}$ follow independently and identically distributed (i.i.d.) Type I Extreme Value (Gumbel) distribution. This assumption yields a closed-form expression for the choice probability:
\begin{equation}
    P_{ni} = \frac{e^{V_{ni}}}{\sum_{j \in \V} e^{V_{nj}}},\label{eq:mnl}
\end{equation}
which is equivalent to applying the softmax function to the deterministic utilities. In the MNL and many ANN-based choice models \citep[e.g.][]{wang2020deepASU} that rely on the i.i.d.\ error assumption, the utility term\footnote{Unless otherwise specified, the term “utility” refers to the deterministic component $V_{ni}$.} is specified as a function only based on the alternative itself, i.e., $V_{ni} = f_i(\mathbf{x}_{ni})$, without considering the dependence between alternatives. However, in spatial choice problems, the deterministic component $V_{ni}$ may exhibit spatial dependencies across alternatives.

Hence we propose GNNs to specify the utility function, where the utility of each alternative is determined not only by its own attributes but also by those of its neighbors in a predefined alternative graph $\G$. This approach absorbs spatial dependencies into the deterministic part of the utility function, analogous to spatial econometric models \citep{lesage2009introduction}. Specifically, in a GNN model with $K_g$ layers, the utility of an alternative $i$ is computed as:
\begin{align}
V_{ni} &= \mathbf{w}^{\top} \mathbf{h}_{ni}^{(K_g)}, \label{eq:projection}\\
\mathbf{h}_{ni}^{(k)} &=\phi^{(k)}\left(\mathbf{h}_{ni}^{(k-1)}, \bigoplus_{j \in \mathcal{N}(i)}^{(k)} \psi^{(k)}\left(\mathbf{h}_{ni}^{(k-1)}, \mathbf{h}_{nj}^{(k-1)}, \mathbf{e}_{n,ij}\right)\right)\, \forall k \in \{1,\ldots,K_g\}, \label{eq:message_passing}
\end{align}
where $\mathbf{h}_{ni}^{(k)}\in \mathbb{R}^{h_k}$ is a vector for the hidden representation of node $i$ in the $k$-th layer of the GNN, with initial representation $\mathbf{h}_{ni}^{(0)}\equiv \mathbf{x}_{ni}$ or using an input embedding function $\mathbf{h}_{ni}^{(0)}= f(\mathbf{x}_{ni})$. The set $\N(i)$ denotes the direct neighbors of node $i$ in the alternative graph $\G$. Each layer of the GNN contain a ``message-passing'' process, where the hidden representation of each node is updated based on its previous representation and the aggregated messages from its neighbors: First, the \textit{message} function $\psi$ encodes the information of node $i$ and its neighbor $j$, along with possible edge features $\mathbf{e}_{n,ij}$ that may represent the weight between nodes $i$ and $j$. Next, the \textit{aggregation} function $\bigoplus$ aggregates these messages from the neighbors of node $i$. Finally, the \textit{update} function $\phi$ updates the node representation based on the aggregated message. After reaching the last GNN layer, the utility is obtained by projecting node representation $\mathbf{h}_{ni}^{(K_g)}$ onto a scalar via a learnable weight vector $\mathbf{w}$ as shown in Eq.~\eqref{eq:projection}. This utility is then used in Eq.~\eqref{eq:mnl} to compute the choice probability.

\begin{figure}[!htbp]
    \centering
    \includegraphics[width=0.8\textwidth]{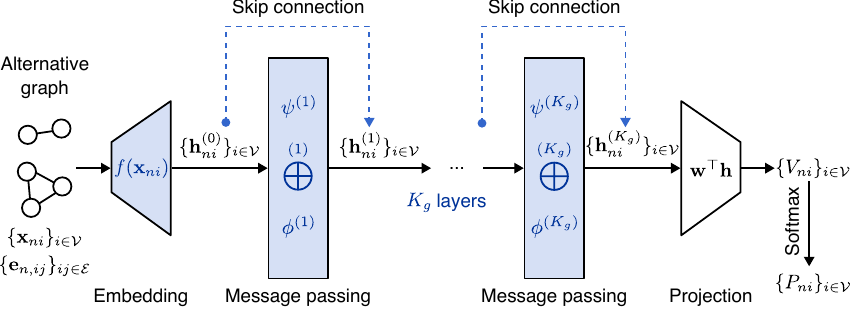}
    \caption{Illustration of the GNN-DCM framework. Customizable design components are colored in blue.}
    \label{fig:gnn_framework}
\end{figure}

Figure.~\ref{fig:gnn_framework} illustrates the GNN-DCM framework. Given the extensive research on GNNs, this framework is highly flexible and can adapt to a wide range of specifications. Such flexibility arises from choices regarding the embedding function, number of layers, message, aggregation, and update functions. The customizable design components in Figure.~\ref{fig:gnn_framework} are colored in blue. For instance, the aggregation function can be a simple sum, mean, or more complicated attention-based sum \citep{velivckovic2018graph}, or any other permutation-invariant operation. In addition, Figure.~\ref{fig:gnn_framework} highlights the use of optional skip connections, which are not part of the standard message-passing formulation in Eq.~\eqref{eq:message_passing} but commonly employed in deep GNNs to alleviate the over-smoothing problem \citep{hamilton2017inductive}. These skip connections allow information to bypass intermediate layers, facilitating more effective gradient flow from earlier to later layers. The implementation details of the skip connection are elaborated in Appendix~\ref{sec:skip_connections}.

For additional techniques in GNN design, such as regularization strategies and multi-relational graphs, we refer readers to \citep{you2020design, hamilton2020graph}. In the next subsection, we demonstrate that the GNN-DCM utility functions generalize classical DCMs and various ANN-based DCMs.

\subsection{GNN-DCM generalizes classical logit models}\label{sec:connection}
The NL model by \cite{mcfadden1978modelling} is the most common member in the GEV family, which captures the correlation between alternatives assuming random components following a GEV distribution. The SCL model by \cite{bhat2004mixed} is a variant of the NL model tailored for residential location choice, where the nests are formed based on spatial adjacency of alternatives. We examine how NL and SCL models are special cases in the GNN-DCM framework, showing the connections between the nested structure and the alternative graph.

\subsubsection{NL is a specific case in GNN-DCMs}
\begin{prop}\label{prop:nl_gnn}
    A two-level nested logit model, where each alternative belongs to a single nest, is a single-layer GNN. In this representation, each nest corresponds to a complete subgraph with a self-loop at each node, and there is no edge between nests.
\end{prop}

\begin{proof}
    In a NL model, denoted by $K_n$ as the total number of nests, and $\mu_k$ as nest-specific independence parameter for nest $B_k$, the choice probability of an alternative $i$ in nest $B_k$ is typically given by:
\begin{align}
    P_{n i} &= P_{ni|B_k}P_{nB_k} \notag \\
    &=\frac{\exp \left(V_{n i}/\mu_k\right)}{\sum_{j \in B_k} \exp \left(V_{n j}/\mu_k\right)} \times \frac{\left(\sum_{j \in B_k} \exp \left(V_{n j}/\mu_k\right)\right)^{\mu_k}}{\sum_{l=1}^{K_n}\left(\sum_{j \in B_l} \exp \left(V_{n j}/\mu_l\right)\right)^{\mu_l}} \notag\\
    &= \frac{\exp \left(V_{n i}/\mu_k\right)\left(\sum_{j \in B_k} \exp \left(V_{n j}/\mu_k\right)\right)^{\mu_k -1}}
    {\sum_{l=1}^{K_n}
    \left(\sum_{j \in B_l} \exp \left(V_{n j}/\mu_l\right)\right)^{\mu_l}
    } \notag \\
    &= \frac{\exp \left(V_{n i}/\mu_k\right)\left(\sum_{j \in B_k} \exp \left(V_{n j}/\mu_k\right)\right)^{\mu_k -1}}
    {\sum_{l=1}^{K_n}
    \left(
    \frac{\sum_{m \in B_l} \exp \left(V_{n m}/\mu_l\right)}{\sum_{m \in B_l} \exp \left(V_{n m}/\mu_l\right)}
    \left(\sum_{j \in B_l} \exp \left(V_{n j}/\mu_l\right)\right)^{\mu_l}
    \right)} \notag \\
    &= \frac{\exp \left(V_{n i}/\mu_k\right)\left(\sum_{j \in B_k} \exp \left(V_{n j}/\mu_k\right)\right)^{\mu_k -1}}
    {\sum_{l=1}^{K_n} \sum_{m \in B_l}
    \left(
    \frac{ \exp \left(V_{n m} / \mu_l\right)}{\sum_{m \in B_l} \exp \left(V_{n m} / \mu_l\right)}
    \left(\sum_{j \in B_l} \exp \left(V_{n j}/\mu_l\right)\right)^{\mu_l}
    \right)} \notag \\
    & = \frac{\exp \left(V_{n i}/\mu_k\right)\left(\sum_{j \in B_k} \exp \left(V_{n j}/\mu_k\right)\right)^{\mu_k -1}}
    {\sum_{l=1}^{K_n} \sum_{m \in B_l}
    \left(
    \exp \left(V_{n m} / \mu_l\right)
    \left(\sum_{j \in B_l} \exp \left(V_{n j}/\mu_l\right)\right)^{\mu_l - 1}
    \right)} \notag\\
    & = \frac{\exp \left(V_{n i}/\mu_k + (\mu_k -1)\log\left( \sum_{j \in B_k} \exp \left(V_{n j}/\mu_k\right)\right)
    \right)}
    {\sum_{l=1}^{K_n} \sum_{m \in B_l}
    \exp \left( V_{n m} / \mu_l + {(\mu_l - 1)
    \log \left(\textstyle\sum_{j \in B_l} \exp \left(V_{n j}/\mu_l \right) \right)} \right)}. \label{eq:nl}
\end{align}

Eq.~\eqref{eq:nl} follows the MNL form given in Eq.~\eqref{eq:mnl}, and the exponential terms in both the numerator and the denominator have the same functional form. We can find that the exponential terms follow the GNN message passing scheme:
\begin{equation}
    V_{ni}^{(1)} = \underbrace{V_{n i}^{(0)} / \mu_k + (\mu_k - 1){\overbrace{
    \log \left(\textstyle\sum_{j \in \N(i) \cup \{i\}} \exp \left(V_{n j}^{(0)}/\mu_k \right) \right)}^{\text{Aggregate } \bigoplus}}}_{\text{Update } \phi},\label{eq:nl_gnn}
\end{equation}
where the aggregate function $\bigoplus$ has a Log-Sum-Exponential (LSE) form, the hidden representation of a node is a scalar, $\mathbf{h}_{ni}^{(k)}=V_{ni}^{(k)}$, and the GNN has a single layer ($K_g=1$). Moreover, this equivalence indicates that the following properties are needed to turn a GNN into a NL model. (1) Self-loops: Message for each nest is aggregated from the set $i \in \mathcal{N}(i) \cup \{i\}$, meaning that each node not only receives messages from its neighbors but also from itself via a self-loop. (2) Complete subgraphs within nests: All alternatives within a nest share the same message with each other, forming a complete subgraph for each nest. (3) No inter-nest edges: Eq.~\eqref{eq:nl} shows that the message for each alternative is only aggregated from its own nest. Thus, there are no edges between alternatives in different nests. (4) Nest-specific parameter: The parameter $\mu_k$ in the nested logit model quantifies the degree of independence among alternatives within a nest.
\end{proof}

\begin{figure}[!ht]
    \centering
    \begin{tabular}{m{0.46\textwidth}|m{0.48\textwidth}}
    \toprule
    The nest structure of alternatives: & The equivalent alternative graph: \\
    \makebox[\linewidth][c]{\includegraphics[width=0.25\textwidth]{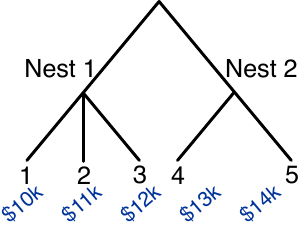}} &
    \makebox[\linewidth][c]{\includegraphics[width=0.25\textwidth]{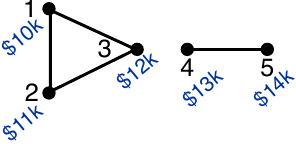}} \\


    The nested logit view to calculate $P_1$:  & The GNN view to calculate $P_1$: \\

    \begin{minipage}{0.5\textwidth}
        \small
        \vspace{0.2cm}

        $P_{B_1}=\frac{\left(e^{\frac{-13}{0.5}}+e^{\frac{-14}{0.5}}\right)^{0.5}}{\left(e^{\frac{-10}{0.6}}+e^{\frac{-11}{0.6}}+e^{\frac{-12}{0.6}}\right)^{0.6}+ \left(e^{\frac{-13}{0.5}}+e^{\frac{-14}{0.5}}\right)^{0.5}}=0.9523$

        \vspace{0.2cm}

        $P_{1|B_1}=\frac{e^{\frac{-13}{0.5}}}{e^{\frac{-13}{0.5}}+e^{\frac{-14}{0.5}}}=0.7307$

        \vspace{0.2cm}

        $P_1 = P_{1|B_1} P_{B_1} = 0.7307 \times 0.9523 = \color{Maroon}{0.6959}$

        \vspace{0.2cm}
        \end{minipage}
        &
        \begin{minipage}{0.5\textwidth}
            \small
            \vspace{0.2cm}

            $V_1=\frac{-10}{0.6} + (0.6-1)\log(e^{\frac{-10}{0.6}}+e^{\frac{-11}{0.6}}+e^{\frac{-12}{0.6}})= -10.063$

            $V_2=\frac{-11}{0.6} + (0.6-1)\log(e^{\frac{-10}{0.6}}+e^{\frac{-11}{0.6}}+e^{\frac{-12}{0.6}})=-11.313$

            $V_3=\frac{-12}{0.6} + (0.6-1)\log(e^{\frac{-10}{0.6}}+e^{\frac{-11}{0.6}}+e^{\frac{-12}{0.6}})=-12.563$

            $V_4=\frac{-13}{0.5} + (0.5-1)\log(e^{\frac{-13}{0.5}}+e^{\frac{-14}{0.5}})=-13.028$

            $V_5=\frac{-14}{0.5} + (0.5-1)\log(e^{\frac{-13}{0.5}}+e^{\frac{-14}{0.5}})=-14.140$

            $P_1 = \frac{e^{V_1}}{e^{V_1}+e^{V_2}+e^{V_3}+e^{V_4}+e^{V_5}} = \color{Maroon}{0.6959}$

            \vspace{0.2cm}
        \end{minipage}
        \\
    \bottomrule
    \end{tabular}
    \caption{Using GNN as the NL model to calculate the choice probability for alternative 1, where the observed utility is defined as $\text{Utility}_i=-\text{Price}_i$; $\mu_1=0.6$, $\mu_2=0.5$. The two approaches yield the same result.}
    \label{fig:nll_gnn_example}
\end{figure}

An example in Figure.~\ref{fig:nll_gnn_example} illustrates the equivalence between an NL model and a single-layer GNN. The toy choice set contains five alternatives grouped into two nests: $B_{1}=\{1,2,3\}$ and $B_{2}=\{4,5\}$ with nest-specific independence parameters $\mu_{1}=0.6$ and $\mu_{2}=0.5$. In the equivalent GNN representation (right panel), every pair of alternatives within the same nest is linked. Each alternative is annotated with a housing price, and the observed utility is defined as $\text{Utility}_i=-\text{Price}_i$. When using the UPDATE rule from Eq.~\eqref{eq:nl_gnn}, the GNN produces exactly the same choice probability for alternative 1, $P_1 = 0.6959$, as the nested logit model. The identical probabilities apply for all other alternatives, thereby confirming Proposition~\ref{prop:nl_gnn}.

\subsubsection{SCL is a specific case in GNN-DCMs}
\label{sec:scl}

SCL model \citep{bhat2004mixed} is a type of NL model tailored for residential location choice, where each nest is a pair of adjacent alternatives (see its structure from the left panel of Figure.~\ref{fig:toy_scl}). The choice probability of the SCL model is:
\begin{equation}
    P_{ni} =\frac{\sum_{j \neq i}\left(\alpha_{i, i j} e^{V_{ni}}\right)^{1 / \mu}\left[\left(\alpha_{i, i j} e^{V_{ni}}\right)^{1 / \mu}+\left(\alpha_{j, ij} e^{V_{nj}}\right)^{1 / \mu}\right]^{\mu-1}}
    {\sum_{k=1}^{\left|\V\right|-1} \sum_{l=k+1}^{\left|\V\right|}\left[\left(\alpha_{k, k l} e^{V_{nk}}\right)^{1 / \mu}+\left(\alpha_{l, k l} e^{V_{nl}}\right)^{1 / \mu}\right]^\mu}, \label{eq:scl}
\end{equation}
where $\alpha_{i, i j}$ is a parameter representing the allocation of alternative $i$ to the paired nest with alternatives $i$ and $j$, there is $0<\alpha_{i,ij}<1$ and $\sum_{j}\alpha_{i,ij}=1$; $\mu$ is the dissimilarity parameter with $0<\mu\le 1$, when $\mu=1$, the SCL model reduces to the MNL model.

\begin{prop}\label{prop:scl_gnn}
    A SCL model is a single-layer GNN, where each nest with a pair of alternatives is an edge in the graph. The message passing scheme takes the following form:
\begin{equation}\label{eq:scl_gnn}
    V_{ni}^{(1)}
    = \log \left( \sum_{j \in \N(i)} \exp \left( V^{(0)}_{ni} + (\mu-1)\log\left[\exp(V^{(0)}_{ni})+\exp(V^{(0)}_{nj})\right]\right)\right),
\end{equation}
where $V^{(0)}_{ni} = f(\mathbf{x}_{ni}) = \frac{\mathbf{b}^{\top}\mathbf{x}_{ni}}{\mu} + \frac{1}{\mu} \log \alpha_{i, ij}$ when using a linear utility function.
\end{prop}

The proof of Proposition~\ref{prop:scl_gnn} is shown in Appendix~\ref{sec:proof_scl_gnn}. Equation \ref{eq:scl_gnn} presents a single-layer GNN, which can be decomposed into a three-step message passing algorithm:
\begin{enumerate}
    \item \textbf{Node embedding.} The node utility value is initialized as $V^{(0)}_{ni} = f(\mathbf{x}_{ni}) = \frac{\mathbf{b}^{\top}\mathbf{x}_{ni}}{\mu} + \frac{1}{\mu} \log \alpha_{i, ij}$, where $\mathbf{b}$ is a vector of coefficients.
    \item \textbf{Message calculation.} The message of a neighboring node $j$ to node $i$ is computed by $M_{i,j} = \psi(V_{ni}^{(0)}, V_{nj}^{(0)}) = V^{(0)}_{ni} + (\mu - 1) \log [\exp (V^{(0)}_{ni}) + \exp (V^{(0)}_{nj})]$.
    \item \textbf{Node aggregation.} Messages from all neighbors of a node $i$ are aggregated by a LSE function: $V^{(1)}_{ni} = \bigoplus (\{M_{i,j}, j \in \N(i)\}) = \log \sum_{j \in \N(i)} \exp (M_{i,j})$.
\end{enumerate}

While here we demonstrate a GNN perspective into SCL models, the message passing algorithm also highlights the unique characteristics of the SCL model, as opposed to the standard GNN family. (1) The passed message in SCL models is a one-dimensional utility value with mostly a linear form. While the initialization function $f(\cdot)$ in standard GNN is often a neural network with a high-dimensional mapping. (2) The LSE aggregation is rarely used in standard GNNs, while it is the standard form for SCL and NL models. (3) The SCL model corresponds to a one-layer GNN since it aggregates the information from the one-hop neighbors of the targeting node $i$. This one-layer GNN is also a unique choice in SCL models because a typical GNN model would adopt around two to three layers to incorporate the information from multi-hop neighborhoods. (4) Lastly, the SCL model shares the parameter $\mu$ across its whole algorithm, as opposed to the standard nested logit model, where each nest has its own $\mu_k$.

\subsection{GNN-DCMs generalize ANN-based choice models}
Beyond the classical NL family, the GNN-DCM framework also generalizes a broad class of ANN-based discrete choice models. This generalization depends on how the attributes of different alternatives are used to construct the utility function of the target alternative. For example, the GNN-DCM framework incorporates the alternative-specific utility functions (ASU-DNN) \citep{wang2020deepASU} as a specific case:
\begin{prop}
The GNN-DCM reduces to the ASU-DNN model \citep{wang2020deepASU} when the embedding function $f(\cdot)$ is a multilayer perceptron (MLP) as in a feedforward neural network, and without message passing, i.e., $\phi(\cdot) = \emptyset, \psi(\cdot) = \emptyset$.
\end{prop}

Essentially, the ASU-DNN model corresponds to a zero-layer GNN, where no information is exchanged between alternatives through message passing over the spatial network. Similarly, a wide range of ANN-based DCMs can be interpreted as special cases of the GNN-DCM framework without message passing, e.g., the TasteNet by \citep{han2022neural}.

The ANN-based DCMs typically capture the alternative interactions in two ways: (1) incorporating the attributes of all alternatives as joint inputs into the ANN-based utility function \citep[e.g.,][]{wang2020deep, wong2021reslogit}; (2) embedding neural networks within the utility function structure of classical logit models \citep[e.g.,][]{sifringer2020enhancing}. These two approaches can both be interpreted within the GNN-DCM framework. The first corresponds to a fully connected GNN, where each alternative is connected to all others, and the message passing functions are implemented using neural networks. The second corresponds to a GNN with neural networks as the input embedding function followed by a message passing scheme defined by Eq.~\eqref{eq:nl_gnn}. Therefore, the GNN-DCM framework unifies both modeling strategies under a single, graph-structured formulation.

\begin{figure}[htbp]
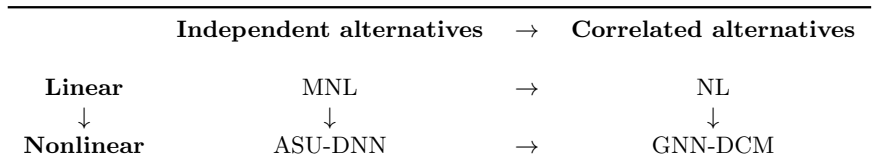

    \centering
    \footnotesize
        \begin{tabular}{cccc}
        \toprule
            & \textbf{Independent alternatives} & $\rightarrow$ & \textbf{Correlated alternatives} \\
\\
        \textbf{Linear}   &   MNL  & $\rightarrow$ &   NL \\
        $\downarrow$ &  $\downarrow$ & & $\downarrow$  \\
        \textbf{Nonlinear}   &   ASU-DNN & $\rightarrow$ &    GNN-DCM   \\
        \bottomrule
        \end{tabular}%
        \caption{The relationship between classical and NN-based discrete choice models.}
        \label{fig:models_connection}
    \vspace{0.2cm}
\end{figure}%

Figure~\ref{fig:models_connection} summarizes the key relationships between the GNN-DCM, existing DCMs, and ANNs. The second row of the figure represents the early development of DCMs, from the MNL model with independent alternatives to the NL model for hierarchically correlated alternatives, and both models use linear utility functions. The second column illustrates recent advances in deep learning-based models that introduce nonlinearity into utility specification. The GNN-DCM framework bridges these two dimensions: it generalizes ANN-based models by enabling flexible alternatives' correlation through message passing, and it extends classical models like the NL and SCL models by allowing richer forms of utility specification via vectorized representations, more expressive message passing scheme, and flexible network designs.

\subsection{Interpreting GNNs for travel demand modeling}\label{sec:substitution}
Previous studies have demonstrated that deep neural networks can provide full economic information for discrete choice models \citep{wang2020deep}. The GNN-DCM framework is no exception. Through numerical methods, we can compute the elasticity, substitution patterns, and visualize the choice probability functions for GNNs. Particularly, substitution patterns describe how the change in an alternative's attributes (e.g., the housing price) affects the choice probabilities of other alternatives. The MNL model is well known for its proportionate substitution pattern: a change in the utility of an alternative proportionally affects the choice probabilities of other alternatives. This behavior is a direct consequence of the IIA property, meaning that the relative choice probabilities between two alternatives are unaffected by the presence or attributes of other alternatives. The IIA property has been criticized for being unrealistic in some applications. For example, in residential location choice models, an increase in housing prices in one community is expected to impact neighboring communities more than distant ones -- the MNL model fails to capture this. We will see that GNN-DCMs mitigate this constraint imposed by the IIA.

Substitution patterns of a choice model can be measured by elasticities $E_{i\, z_{nj}}$, which is defined by the percentage change in the choice probability of individual $n$ on alternative $i$ in response to a percentage change in the attribute of alternative $j$:
\begin{equation*}
    E_{i, z_{nj}}=\frac{\partial P_{ni}}{\partial z_{nj}} \frac{z_{nj}}{P_{ni}},
\end{equation*}
where $z_{nj}$ is the attribute of alternative $j$ that is being changed. When $i=j$, it is called direct elasticity; when $i\neq j$, it is called cross-elasticity. The cross-elasticities of the MNL model are the same for all $i\neq j$.

We investigate the substitution patterns of GNN-DCMs by deriving their elasticities. Eq.~\eqref{eq:projection} and Eq.~\eqref{eq:message_passing} show that $V_j$ is a function of the attributes of alternative $j$ and its $k_g$-hop neighbors, $\mathcal{N}_{k_g}(j)$, but not of other alternatives. This leads to the following results:
\begin{itemize}
\item For an alternative $i$ that is not a $k_g$-hop neighbors of $j$, i.e. $i \notin \mathcal{N}_{k_g}(j)\cup \{j\}$, the cross-elasticity in response to the change in the attribute $z_{nj}$ of alternative $j$ is:
\begin{align*}
    E_{i\, z_{nj}} &= \frac{\partial \left( e^{V_{ni}} /\sum_{k \in \V} e^{V_{nk}}\right)}{\partial z_{nj}} \frac{z_{nj}}{P_{ni}} \\
    &= \frac{-e^{V_{ni}}\left(\sum_{k\in {\mathcal{N}_{k_g}(j) \cup j}}e^{V_{nk}}\frac{\partial V_{nk}}{\partial z_{nj}}\right)}{\left(\sum_{k\in \mathcal{V}} e^{V_{nk}}\right)^2}\frac{z_{nj}}{P_{ni}}\\
    &=-P_{ni} \sum_{k\in {\mathcal{N}_{k_g}(j) \cup j}}P_{nk}\frac{\partial V_{nk}}{\partial z_{nj}}\frac{z_{nj}}{P_{ni}}\\
    &=-z_{nj}\sum_{k\in {\mathcal{N}_{k_g}(j) \cup j}}P_{nk}\frac{\partial V_{nk}}{\partial z_{nj}},
\end{align*}
which is irrelevant to the alternative $i$. This means that the cross-elasticities are the same for all alternatives outside of the $k_g$-hop neighbors of $j$, the same as the proportionate substitution pattern of the MNL model.
\item For an alternative $j$ belongs to the $k_g$-hop neighbors of $j$, i.e., $i \in \mathcal{N}_{k_g}(j)\cup \{j\}$, the (cross-)elasticity takes the form
\begin{align*}
    E_{i\, z_{nj}} &= \frac{e^{V_{ni}}\frac{\partial e^{V_{ni}}}{\partial z_{nj}}\sum_{k\in \mathcal{V}} e^{V_{nk}} -e^{V_{ni}}\left(\sum_{k\in {\mathcal{N}_{k_g}(j) \cup j}}e^{V_{nk}}\frac{\partial V_{nk}}{\partial z_{nj}}\right)}{\left(\sum_{k\in \mathcal{V}} e^{V_{nk}}\right)^2}\frac{z_{nj}}{P_{ni}}\\
    &=\left(P_{ni}\frac{\partial e^{V_{ni}}}{\partial z_{nj}} -P_{ni} \sum_{k\in {\mathcal{N}_{k_g}(j) \cup j}}P_{nk}\frac{\partial V_{nk}}{\partial z_{nj}} \right)\frac{z_{nj}}{P_{ni}}\\
    &=\left(\frac{\partial e^{V_{ni}}}{\partial z_{nj}} - \sum_{k\in {\mathcal{N}_{k_g}(j) \cup j}}P_{nk}\frac{\partial V_{nk}}{\partial z_{nj}} \right)z_{nj},
\end{align*}
which varies across alternatives $i$. This means that the cross-elasticities of alternatives within the $k_g$-hop neighbors of $j$ are alternative-specific and depend on the attributes of $i$, $j$ and its $k_g$-hop neighbors. This is a key difference from the MNL model.
\end{itemize}

These results demonstrate that GNN-DCMs allow for more flexible substitution patterns, where a change in an alternative's attributes has dedicated localized effects on $k_g$-hop neighbors, rather than globally proportionate as in the MNL model. Since NL and SCL models are specific one-layer GNN-DCMs, deeper GNN models also present more flexible elasticities than NL and SCL models. By choosing the number of layers (i.e., $k_g$) and constructing different graph topologies, we can control the behavior of substitution patterns between alternatives. We name it spatially-aware substitution patterns, since the substitution patterns depend on the spatial relationships defined by the graph structure. An illustration of this behavior is shown in Section~\ref{sec:exp_elasticities}.

\section{Experiment Setup}\label{sec:case study}

\subsection{Dataset}
\begin{figure}[!ht]
    \centering
    \begin{subfigure}{0.4\textwidth}
        \centering
        \includegraphics[width=\linewidth]{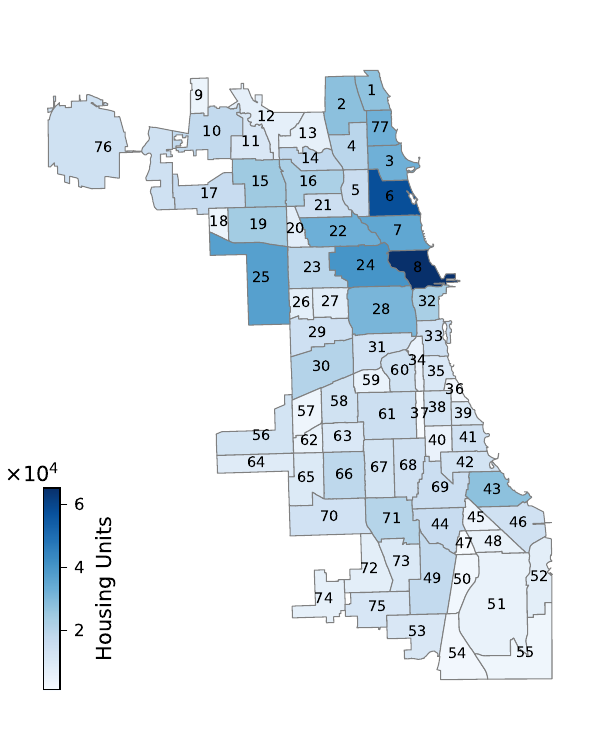}
        \caption{The 77 communities in Chicago.}
        \label{fig:communities_map}
    \end{subfigure}
    \hspace{0.0\linewidth} 
    \begin{subfigure}{0.4\textwidth}
        \centering
        \includegraphics[width=\linewidth]{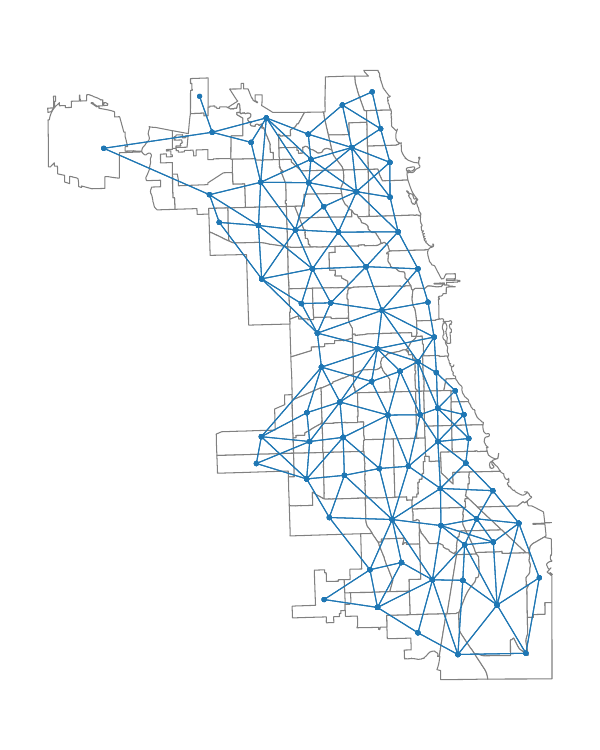}
        \caption{The graph structure.}
        \label{fig:communities_graph}
    \end{subfigure}
    \caption{The study area and the alternative graph.}
    \label{fig:communities}
\end{figure}

We perform a case study on residential location choice across 77 communities in Chicago. As of the 2020 census, the study area has a population of approximately 2.7 million. The locations of these communities and the number of housing units in each are illustrated in Figure.~\ref{fig:communities_map}. The 77 community areas were originally defined in the 1920s by the University of Chicago Social Science Research Committee and were intended to represent moderately coherent social characteristics across urban space at a generalized geographical scale. Although the geography and population of these communities have evolved over time, their boundaries have remained relatively stable. These community areas continue to provide a consistent framework for urban planning, statistical analysis, and policy-making \citep{chicago_communities}. The Chicago Metropolitan Agency for Planning (CMAP) and other agencies have adopted these community areas as a standard for reporting demographic and socioeconomic data.

The goal of this study is to investigate how households select their residential locations within these communities. The social and demographic characteristics of the communities are derived from census data obtained from the American Community Survey 5-year data (2013-2017)\footnote{\url{https://www.nhgis.org/}} and land use data (2017) from the CMAP Data Hub\footnotemark[2]. Household-level information, such as home location and household income, is sourced from the My Daily Travel Survey data\footnotetext[2]{\url{https://datahub.cmap.illinois.gov/}}, conducted by CMAP (released in 2019 with data collected in 2017). The travel survey captures the travel patterns of over 12,000 households in the seven-county region of northeastern Illinois. For this study, we extracted a subset of 3,838 households residing in these 77 communities who reported their work locations. Although the sample size is relatively small compared to the total population of these communities, it serves as a convenient dataset for our analysis and is sufficiently large when compared to similar studies in the literature, as indicated in Table~\ref{tab:dataset}.

\begin{table}[!htbp]
    \centering
    \small
    \caption{Dataset in residential location choice studies.}
      \begin{tabular}{lllll}
      \toprule
      Paper & Methods & City  & \# Zones & \# Households \\
      \midrule
      \citep{bhat2004mixed} & SCL   & Dallas & 98 & 236 \\
      \citep{sener2011accommodating} & GSCL  & San Francisco & 115 & 702 \\
      \citep{perez2022spatially} & SCNL  & Santander (Spain) & 26 & 534 \\
      \textbf{This work}  & GNN   & Chicago & 77 & 3838 \\
      \bottomrule
      \end{tabular}%
    \label{tab:dataset}%
\end{table}%

After reviewing the literature \citep{bhat2004mixed, sener2011accommodating, schirmer2014role}, considering data availability, and conducting significance tests in the MNL model, we selected thirteen attributes for modeling household residential choice behavior, as detailed in Table~\ref{tab:attributes}. These variables are categorized into four groups: housing, land use, transportation, and demographics. All attributes are scaled to ensure comparable magnitudes across variables. When household members have different work locations, we use the furthest distance as the ``work distance'' attribute. Although the choices of work and home locations are mutually affected, we treat work locations as exogenous in this study, as the focus is on residential location choice. It is worth noting that directly incorporating household-level attributes, such as household income, into the utility function of the MNL model does not affect the choice probability, as the MNL model relies on relative differences in utility. Therefore, we interact household-level attributes with community-level attributes, creating attributes ``black interact'', ``white interact'', and ``income interact'', as shown in Table~\ref{tab:attributes}. This approach aligns with common practices in the literature \citep{bhat2004mixed, sener2011accommodating, perez2022spatially}.

It is also noteworthy that GNNs can directly utilize household-level attributes as node features, as the projection matrices in Eq.\eqref{eq:projection} and Eq.\eqref{eq:message_passing} inherently interact household-level with community-level attributes. This eliminates the need for manually constructing interactive terms and allows for the inclusion of additional household-level attributes—such as vehicle ownership or the presence of children—that are challenging to incorporate into the MNL model. However, to maintain a fair comparison, we retain the same set of attributes for the GNNs as those used in the MNL model. This also enables us to focus on differences in model structures rather than variations in input features.

\begin{table}[htbp]
  \centering
  \footnotesize
  \begin{threeparttable}
  \caption{Summary of Variables.}
    \begin{tabular}{lp{25em}l}
    \toprule
    Categories & Attributes & Short name \\
    \midrule
    \multirow{3}{*}{Housing} & Number of housing units in the community\tnote{*} &  \# Units \\
    \addlinespace
          & Median housing value in the community\tnote{*} &  House value\\
          \addlinespace
          & Median house age in the community\tnote{*} &  House age\\
    \midrule
    \multirow{5}{*}{Land use} & Land use mixture measured by entropy &  Land mixture\\
    \addlinespace
          & Percentage of land allocated to single-family housing &  \% Single house\\
          \addlinespace
          & Percentage of land allocated to multi-family housing &  \% Multi house\\
          \addlinespace
          & Percentage of land allocated to office &  \% Office \\
    \midrule
    \multirow{2}{*}{Transportation} & Transit accessibility in the community (released by CMAP)\tnote{*} &  Transit access\\
    \addlinespace
          & Log of distance to work &  Work distance\\
          \midrule
          \multirow{4}*{Demographics} & Population density of the community\tnote{*} & Pop density \\
          \addlinespace
          & Percentage of the black population in the community multiplied with the black household dummy variable & Black interact \\
          \addlinespace
        & Percentage of the white population in the community multiplied with the white household dummy variable & White interact \\
        \addlinespace
        & The difference between the household income with the community median household income\tnote{*} & Income interact \\
    \bottomrule
    \end{tabular}%
  \label{tab:attributes}%
  \begin{tablenotes}
    \item[*] Scaled to have comparable magnitude across attributes.
\end{tablenotes}
\end{threeparttable}
\end{table}%

Finally, we use spatial adjacency--the most straightforward approach--to construct the graph structure for the 77 communities. Each community is represented as a node in the graph, with edges connecting adjacent communities. The resulting graph structure is illustrated in Figure.~\ref{fig:communities_graph}.

\subsection{Experiment design}

The experiments are designed to address the following research questions:
\begin{itemize}
    \item \textbf{RQ1:} How does the performance of GNN-DCM compare to the MNL, SCL, and ASU-DNN models?
    \item \textbf{RQ2:} How do different components of GNN design affect the performance of GNN-DCM?
    \item \textbf{RQ3:} How do GNN-DCMs compare with traditional models in terms of interpretability and elasticities?
\end{itemize}

To address \textbf{RQ1}, we compare the GNN-DCM with the standard MNL model and SCL model \citep{bhat2004mixed}. The MNL model assumes independence of irrelevant alternatives, and the SCL model can be viewed as a single-layer GNN with a specific update function defined in Eq.~\eqref{eq:scl_gnn}. We also include a comparison with the ASU-DNN model \citep{wang2020deep}, which serves as a benchmark representing GNNs without any message passing between alternatives. Comparative results are reported in Section~\ref{sec:comparison}.

To address \textbf{RQ2}, we evaluate three representative GNN update methods: Message Passing Neural Network (MPNN), Graph Convolutional Networks (GCN) \citep{kipf2016semi}, and Graph Attention Network (GAT) \citep{velivckovic2018graph}. For MPNN, we also explore different aggregation functions, including sum, max, mean, and log-sum-exp (LSE), the latter being inspired by the SCL model. We use the same number of hidden dimensions for all layers. Our base configuration is a two-layer GNN with 64 hidden dimensions and skip connections. We then systematically vary the number of layers, hidden dimensions, and the use of skip connections. Details of these GNN update methods are provided in Appendix~\ref{sec:model_details}, details of the skip connection can be found in Appendix~\ref{sec:skip_connections}, and the experimental results are presented in Section~\ref{sec:ablation}.

To address \textbf{RQ3}, we assess the interpretability of the GNN-DCM by analyzing how predicted choice probabilities respond to changes in alternative attributes. We also compute the cross-elasticities of the GNN-DCM and compare them with those from the MNL and SCL models. The results are presented in Section~\ref{sec:explanation_choice} and Section~\ref{sec:exp_elasticities}.

\subsection{Implementation details}
Model parameters are estimated by minimizing the negative log-likelihood (NLL) of the training data:
\begin{align*}
    \mathcal{L}(\bm{\theta}) = -\sum_{n=1}^{N} \log P_{ni},
\end{align*}
where $P_{ni}$ denotes the predicted choice probability of household $n$ selecting alternative $i$, and $\bm{\theta}$ represents the model parameters. The MNL and SCL models are trained using the NLL computed over the entire training dataset, where $N$ is the total number of households. In contrast, DNN and GNN models are trained using mini-batch gradient descent, where the NLL is calculated for each batch and used to update parameters iteratively. We use a mini-batch size of 32, a common choice in the deep learning literature \citep{goodfellow2016deep}.

To mitigate randomness during training, we conduct ten-fold cross-validation and report the average performance across all folds. Each fold the dataset is partitioned into a training set (90\%) and a held-out test set (10\%). All models are implemented in PyTorch \citep{Ansel_PyTorch_2_Faster_2024}. The GNN models are built using the PyTorch Geometric library \citep{fey2019fast}, which provides efficient implementations of various GNN layers. The MNL and SCL models are optimized using the L-BFGS optimizer \citep{liu1989limited}, while the GNN and ASU-DNN models are trained with the Adam optimizer \citep{kingma2014adam} using a learning rate of 0.01. Dropout regularization with a rate of 0.05 is applied to prevent overfitting. It takes around 1 minute for training a GNN-DCM at a MacBook Pro with an M3 Pro chip and 18GB of RAM. Additional implementation details are available in the code repository.

\section{Results}
We compare the GNN-DCMs with a set of baseline models using a case study of residential location choice in Chicago. This section showcases the performance, interpretability, and elasticities of the GNN model. The code and data for this study are available at \url{https://github.com/chengzhanhong/GNN_residential_choice} (will be released upon acceptance).

\subsection{Comparing GNN-DCMs to benchmark models}\label{sec:comparison}
We first compare the GNN-DCM model with the highest performance to the benchmark models, including MNL, SCL, and ASU-DNN. The GNN-DCM achieves the highest performance when it uses a GAT convolution layer, 64 latent dimensions, and two layers. For a fair comparison, the ASU-DNN uses the same specification, except that it does not have the message passing component. We also present the performance of GNN with one, two, and three layers, thus revealing the impacts of GNN depth. The average performance of the ten-fold cross-validation of the models is summarized in Table~\ref{tab:performance}. Evaluation metrics include log-likelihood, prediction accuracy, top-5 accuracy, the average distance between the centroids of the predicted and actual communities (Avg. distance), F1 score, and mean reciprocal rank (MRR). The best performance is highlighted in bold, and the second-best performance is underlined and italicized.

\begin{table}[htbp]
    \centering
    \small
    \begin{threeparttable}
    \caption{The average performance of ten-fold cross-validation.}
      \begin{tabular}{lcccccc}
      \toprule
                  & MNL & SCL & ASU-DNN & GNN (1 layer) & GNN (2 layers) & GNN (3 layers) \\
      \midrule
      Log-likelihood & -1342.05 & -1342.05 & -1319.00 & -1308.47 & \underline{\textit{-1306.68}} & \textbf{-1309.41} \\
      Accuracy & 12.14\% & 12.14\% & 12.71\% & 12.64\% & \underline{\textit{13.03\%}} & \textbf{13.26}\% \\
      Top-5 accuracy & 36.37\% & 36.37\% & 37.52\% & 37.78\% & \textbf{38.75\%} & \underline{\textit{38.15\%}} \\
      Avg. distance [km] & 7.38 & 7.38 & 7.43 & 7.01 & \textbf{6.97} & \underline{\textit{7.01}} \\
      F1 score & 0.033 & 0.033 & 0.045 & 0.065 & \underline{\textit{0.066}} & \textbf{0.068} \\
      Mean reciprocal rank & 0.250 & 0.250 & 0.258 & 0.279 & \textbf{0.285} & \underline{\textit{0.284}} \\
      \bottomrule
      \end{tabular}%
    \label{tab:performance}%
    \begin{tablenotes}
    \item[*] In each row, \textbf{the best} number is highlighted in bold, and \underline{\textit{the second-best}} number is underlined and italicized.
    \end{tablenotes}
    \end{threeparttable}
\end{table}

Both neural network-based models (ASU-DNN and GNNs) outperform the classical MNL and SCL models, highlighting the enhanced capability of using non-linear utility functions and attribute interactions. This finding validate the past research that overall the machine learning and deep learning models outperform the classical econometrics models. Interestingly, the ASU-DNN model yields the largest average distance between predicted and actual communities—even higher than the MNL model—suggesting that its lack of spatial structure limits its ability to capture spatial dependencies.

The GNN model consistently outperforms the ASU-DNN, MNL, and SCL models across all evaluation metrics, with the one exception of accuracy in the one-layer GNN configuration. This demonstrates the effectiveness of incorporating spatial dependencies in modeling residential location choices. Among the GNNs, the two-layer and three-layer models perform better than the one-layer model, indicating that deeper architectures can capture more complex interactions among alternatives. However, the three-layer GNN does not substantially outperform the two-layer version and performs slightly worse in top-5 accuracy and mean reciprocal rank, suggesting that adding more layers may not always lead to better performance in this context. This finding could be caused by the oversmoothing issue in GNN. While deeper GNN architecture can pass more messages to the targeting alternative node, deeper architecture also diminish the model's learning capability of differentiating across nodes.

Note that the SCL and the MNL models have the same performance because the estimation value of the dissimilarity parameter approaches one, making the SCL model equivalent to the MNL model. This observation is consistent with findings reported by \cite{sener2011accommodating}. A plausible explanation for this behavior is the SCL model's use of equally distributed allocation parameters across all neighbors and a single dissimilarity parameter for all alternatives. Such an approach may fail to capture the heterogeneous dependencies in some data sets.

\subsection{Impact of GNN-DCM configurations}\label{sec:ablation}
\begin{figure}[!ht]
\begin{center}
\includegraphics[width=\textwidth]{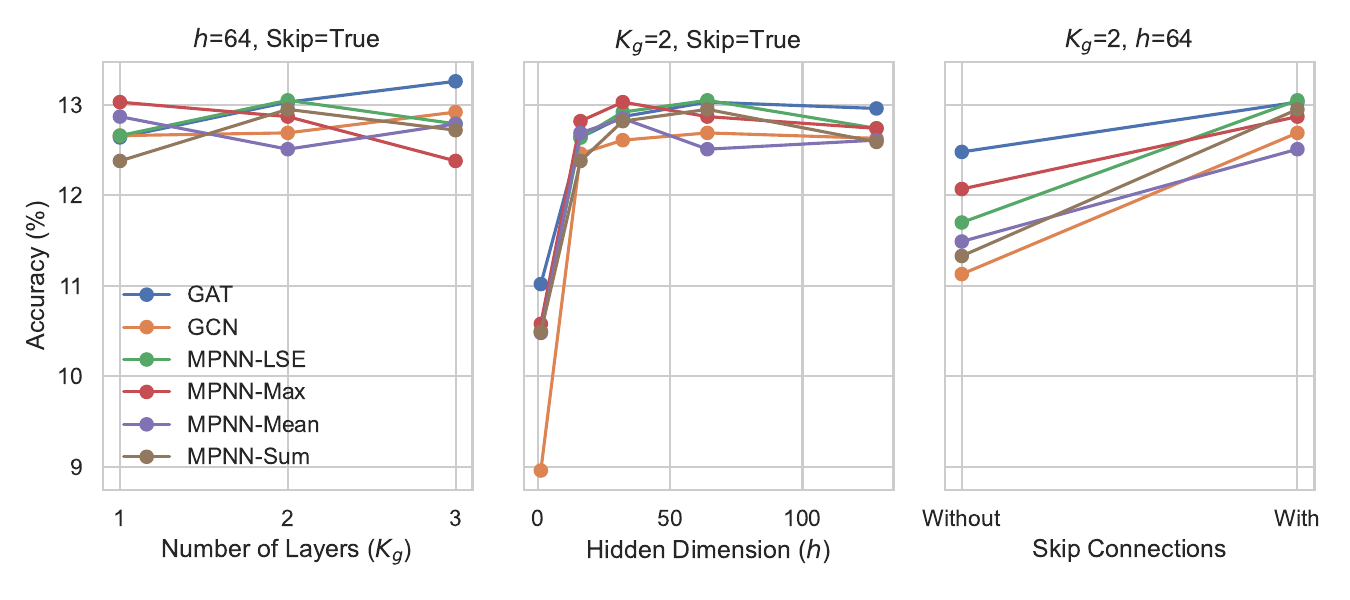}
\caption{Visualization of forecast accuracy of ten-fold cross-validation for different GNN designs.}\label{fig:ablation}
\end{center}
\end{figure}

We then investigate which configuration in the GNN-DCMs contributes the most to their predictive performance. Table~\ref{tab:ablation} presents the average accuracy from ten-fold cross-validation across various GNN configurations, including different update and aggregation functions, hidden dimensions ($h$), number of graph convolution layers ($K_g$), and the use of skip connections. Figure.~\ref{fig:ablation} further visualizes these results.

\begin{table}[!ht]
  \centering
  \small
  \begin{threeparttable}
  \caption{The average accuracy of ten-fold cross-validation for various GNN-DCM designs.}
    \begin{tabular}{cc|ccc|cccc|c}
    \toprule
    \multicolumn{2}{c}{Models} & \multicolumn{3}{c}{$h=64$} & \multicolumn{4}{c}{$K_g=2$} & $h=64, K_g=2$\\
    \midrule
    Name  & $\bigoplus$ & $K_g=1$ & $K_g=2$ & $K_g=3$ & $h=1$ & $h=16$ & $h=32$ & $h=128$ & W/o skip \\
    \midrule
    \multirow{4}[1]{*}{MPNN} & Sum   & 12.38\% & \textbf{12.95\%} & \underline{\textit{12.72\%}} & 10.49\% & 12.38\% & 12.82\% & 12.59\% & 11.33\% \\
          & Max   & \textbf{13.03\%} & 12.87\% & 12.38\% & 10.58\% & 12.82\% & \underline{\textit{13.03\%}} & 12.74\% & 12.07\% \\
          & Mean  & \textbf{12.87\%} & 12.51\% & 12.79\% & 10.48\% & 12.69\% & \underline{\textit{12.85\%}} & 12.61\% & 11.49\% \\
          & LSE   & 12.66\% & \textbf{13.05\%} & 12.79\% & 10.50\% & 12.64\% & \underline{\textit{12.92\%}} & 12.74\% & 11.70\% \\
    \midrule
    GCN   & Sum  & 12.66\% & \underline{\textit{12.69\%}} & \textbf{12.92\%} & 8.96\% & 12.46\% & 12.61\% & 12.63\% & 11.13\% \\
    \midrule
    GAT   & Sum & 12.64\% & \underline{\textit{13.03\%}} & \textbf{13.26\%} & 11.02\% & 12.64\% & 12.87\% & 12.96\% & 12.48\% \\
    \bottomrule
    \end{tabular}%
  \label{tab:ablation}%
      \begin{tablenotes}
        \item[*] In each row, \textbf{the best} number is highlighted in bold, and \underline{\textit{the second-best}} number is underlined and italicized.
    \end{tablenotes}
    \end{threeparttable}
\end{table}%

Overall, the results suggest that model performance is relatively insensitive to the number of layers ($K_g$), but highly sensitive to the hidden dimension. Notably, all the best-performing configurations across models (Message Passing, GCN, and GAT) share a hidden dimension of $h=64$. The middle panel of Figure.~\ref{fig:ablation} illustrates this finding, showing that the performance of all models improves significantly when the hidden dimension increases from 1 to 64, but remains relatively stable for larger dimensions. This indicates that a hidden dimension of 64 is sufficient to capture the complexity of the residential location choice problem, while larger dimensions do not yield substantial performance gains. These results highlight the importance of utilizing high-dimensional representations to capture the complexity of residential location choices.

We further compared the LSE aggregation to other aggregation methods because the LSE aggregation is the most common practice in the classical choice modeling. It also has an appealing utility interpretation in NL models because the log-sum form indicates the utility of the alternatives in a nest. However, the predictive differences across various aggregation functions appear minor. As suggested by Figure 6a, the LSE aggregation in a message passing network achieves comparable performance to other aggregation approaches, including max, mean, and sum. This result suggests that other aggregation methods are as effective as the LSE aggregation in capturing spatial dependencies, as least for predicting residential location choices.

Finally, skip connections consistently lead to higher accuracy across all configurations. When skip connections are removed (“W/o skip” column), the performance of each model drops substantially. This is because skip connections maintain the alternative's original representation after each layer, preventing the node's information from being replaced by the aggregated information from its neighbors. This finding resonates with the past results from both the machine learning and the choice modeling communities. For example, \cite{wang_theory-based_2021} and \cite{wong2021reslogit} both designed a structure similar to ResNet by integrating the MNL outputs with neural networks, indicating the importance of skipping the deeper architecture and retaining the linear components in models.

\subsection{Understanding the behavior of residential location choice}\label{sec:explanation_choice}
It is also essential to understand how various factors influence residential location choices, thus enhancing model interpretation and informing decision-making for transportation policy and planning. Classical MNL and NL models use linear utility functions, where the coefficients directly reflect an attribute's global effect on choice probabilities. This characteristic enhances the interpretability of these models, making them valuable tools in analysis and practice. Table~\ref{tab:MNL_coefficients} shows the estimated coefficients of the MNL model from one of the ten-fold cross-validation experiments. The results generally align with intuitive expectations: households are more likely to choose communities with more housing units, lower house values, and higher population densities. The negative coefficient for distance to work indicates that households prefer to reside closer to their workplaces. Additionally, the positive coefficients for the black and white interaction terms indicate that households are more likely to select communities with a higher percentage of residents matching their own race. Our results show a negative coefficient of the transit accessibility, which has also been observed in previous studies \citep{hu2019housing}; although the effect of accessibility in household residential location choice may have different answers depending on contexts \citep{zondag2005influence,chen2008accessibility}.

\begin{table}[htbp]
    \centering
    \small
    \caption{Estimation results of the MNL model.}
      \begin{tabular}{ccc|ccc}
      \toprule
      Attribute name & Parameter & t-statistic & Attribute name & Parameter & t-statistic \\
      \midrule
      \# Units & 1.972 & 18.85 & Transit access & -1.473 & -8.09\\
      House value & -0.322 & -2.90 & Work distance & -0.908 & -21.63\\
      House age & 0.782 & 7.23 & Pop density & 2.341 & 17.12\\
      Land mixture & -0.944 & -896 & Black interact & 4.001 & 20.02\\
      \% Single house & -0.802 & -5.65 & White interact & 2.812 & 12.34\\
      \% Multi house & 0.581 & 7.63 & Income interact & -0.875 & -5.70\\
      \% Office & 0.107 & 5.48 \\
      \bottomrule
      \end{tabular}%
    \label{tab:MNL_coefficients}%
\end{table}%

\begin{figure}[!htbp]
    \centering
    \begin{subfigure}{0.45\textwidth}
        \centering
        \includegraphics[width=\linewidth]{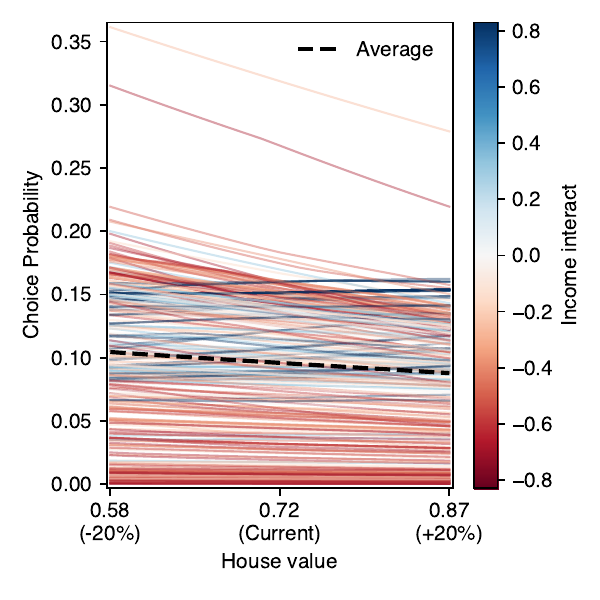}
        \caption{ICE plot of choosing Lake View (community 6) by housing units median value.}
        \label{fig:lake_view_house_value}
    \end{subfigure}
    \hspace{0.05\linewidth} 
    \begin{subfigure}{0.45\textwidth}
        \centering
        \includegraphics[width=\linewidth]{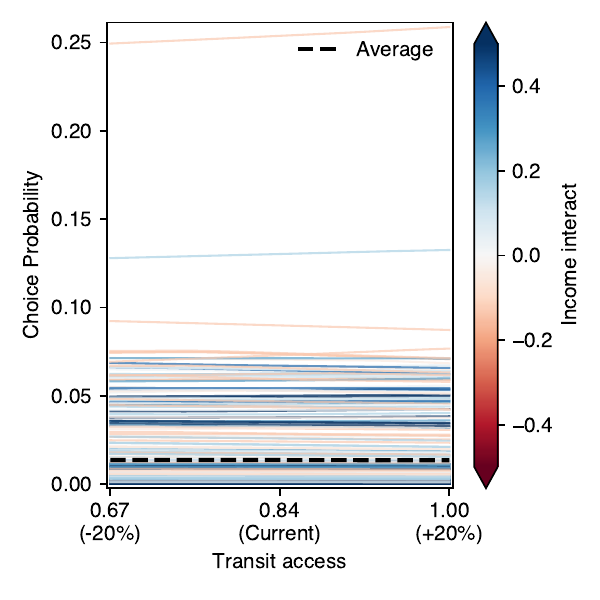}
        \caption{ICE plot of choosing South Shore (community 43) by transit accessibility.}
        \label{fig:south_shore_transit}
    \end{subfigure}
    \caption{The impact of housing unit value and transit accessibility on the choice probability.}
    \label{fig:ICE}
\end{figure}

Neural network-based choice models capture the interactions between attributes in a non-linear way. As a result, the influence of an attribute (e.g., house value) varies with households' attributes, reflecting individuals heterogeneity in their choice behavior. While the deep learning approach improves predictive accuracy, the absence of single interpretable coefficients poses challenges for model interpretation. To overcome this limitation, we employ Individual Conditional Expectation (ICE) plots \citep{goldstein2015peeking} to understand how the GNNs capture choice behavior. Since the ICE plots are alternative- and attribute-specific, we focus on two communities—Lake View (community 6) and South Shore (community 43)—to examine the effects of housing units' median value and transit accessibility on choice probabilities, as shown in Figure.~\ref{fig:ICE}. These communities were selected due to their high respondent counts, representing the entire region and the southern area, respectively.

Each ICE plot displays one line per individual from the test set to show how the choice probability of an individual changes as an attribute changes. To highlight interactions between attributes, we color the lines in Figure.~\ref{fig:ICE} based on the income interaction attribute. Figure.~\ref{fig:lake_view_house_value} reveals that the probability of choosing Lake View generally decreases as the median value of housing units rises, consistent with the MNL model's findings. Interestingly, this effect varies across households with different income levels: high-income households (blue curves) are less sensitive to house value and even show a slight increase in choice probability with higher house value. This suggests that high-income households may prioritize community amenities, social status, or potential capital gains over housing costs. Figure.~\ref{fig:south_shore_transit} shows that most households' choices of living in South Shore remain largely unaffected by transit accessibility, with only a few showing increased probability as accessibility improves. Combined with the MNL results, this implies that transit accessibility is not a primary consideration for most households when selecting South Shore as a residence.

In summary, the GNN models, while lacking single coefficients, can reflect nuanced attribute interactions and household heterogeneity, yielding a more detailed understanding of how factors influence specific alternatives and households. We acknowledge that the MNL model provides straightforward interpretability through its global attribute effects, though its results may be confounded by attribute correlations or non-linear relationships. Results based on GNN offer valuable insights into how the preference of a certain group of households shifts in response to changes in specific factors.

\subsection{Analysis of elasticities}\label{sec:exp_elasticities}
We also investigate how choice probabilities respond to changes in community attributes and how these responses differ between the MNL and the GNN-DCM. To do so, we randomly select a household and calculate the direct and cross elasticities of its choice probability w.r.t. changes in the attributes of Lake View (community 6). The elasticities are computed for both the MNL and GNN (2 layers) models, as shown in Table~\ref{tab:elasticities}.

\begin{table}[htbp]
    \centering
    \small
    \caption{Elasticities of a selected household on community 6 (Lake View).}
    \begin{tabular}{p{3cm} *{6}{>{\centering\arraybackslash}p{1.8cm}}}
        \toprule
        Attribute name & \multicolumn{3}{c}{MNL} & \multicolumn{3}{c}{GNN (2 layers)} \\
        \cmidrule{2-7}
        & \makecell[l]{Direct \\ elasticity} & \makecell[l]{W.r.t. a \\ neighbor} & \makecell[l]{W.r.t. a \\ non-neighbor} & \makecell[l]{Direct \\ elasticity} & \makecell[l]{W.r.t. a \\ neighbor} & \makecell[l]{W.r.t. a \\ non-neighbor} \\
        \midrule
        \# Units & 1.114 & -0.630 & -0.630 & 0.609 & -0.392 & -0.244 \\
        House value & -0.149 & 0.084 & 0.084 & -0.648 & 0.333 & 0.321 \\
        House age & 0.611 & -0.346 & -0.346 & 0.856 & -0.450 & -0.389 \\
        Land mixture & -0.403 & 0.228 & 0.228 & -0.146 & -0.017 & 0.121 \\
        \% Single house & -0.057 & 0.032 & 0.032 & 0.002 & -0.004 & 0.000 \\
        \% Multi house & 0.117 & -0.066 & -0.066 & 0.250 & -0.142 & -0.110 \\
        \% Office & 0.000 & 0.000 & 0.000 & -0.004 & 0.001 & 0.003 \\
        Transit access & -0.910 & 0.514 & 0.514 & 0.098 & -0.130 & -0.011 \\
        Work distance & -0.191 & 0.108 & 0.108 & -0.290 & 0.275 & 0.088 \\
        Pop density & 1.859 & -1.051 & -1.051 & 0.411 & -0.074 & -0.210 \\
        Black pop & 0.000 & 0.000 & 0.000 & 0.000 & 0.000 & 0.000 \\
        White pop & 1.542 & -0.872 & -0.872 & 0.742 & -0.189 & -0.413 \\
        HH income & 0.162 & -0.091 & -0.091 & 0.138 & -0.011 & -0.094 \\
        \bottomrule
    \end{tabular}
    \label{tab:elasticities}
\end{table}

The signs of elasticities are generally consistent across the two models, suggesting that they capture similar choice behaviors for the selected household. For example, both models have positive direct elasticities for the number of housing units, indicating that an increase in the number of housing units in Lake View will lead to a higher probability of choosing Lake View itself. Conversely, the negative cross-elasticities w.r.t. neighboring and non-neighboring communities imply that such an increase reduces the likelihood of choosing other communities. The MNL model, because of the proportionate substitution patterns, the cross-elasticities w.r.t. a neighbor and a non-neighbor are the same for each attribute. In contrast, the GNN model exhibits varying cross-elasticities for neighbors versus non-neighbors, reflecting heterogeneous substitution patterns that account for spatial dependencies.

\begin{figure}[!ht]
    \centering
    \begin{subfigure}{0.4\textwidth}
        \centering
        \includegraphics[width=\linewidth]{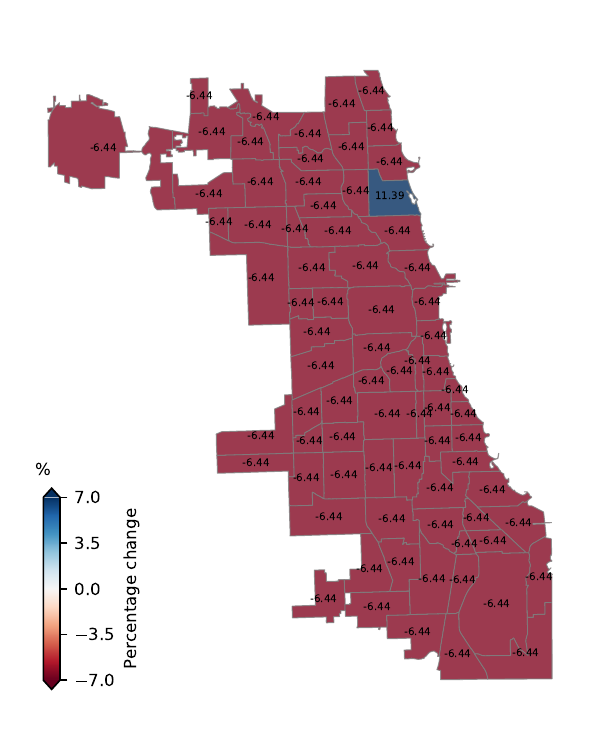}
        \caption{MNL.}
        \label{fig:elasticities_mnl}
    \end{subfigure}
    \hspace{0.0\linewidth} 
    \begin{subfigure}{0.4\textwidth}
        \centering
        \includegraphics[width=\linewidth]{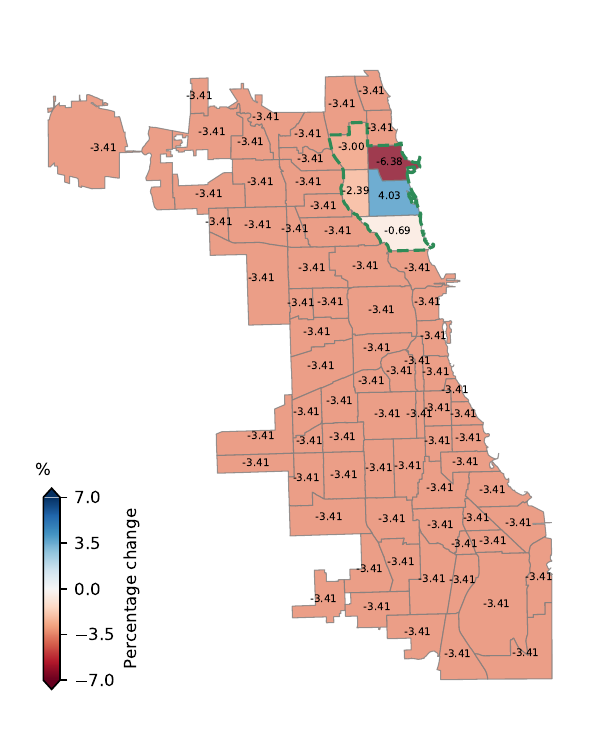}
        \caption{GNN (1 layer).}
        \label{fig:elasticities_gnn1}
    \end{subfigure}

    \vspace{0.1cm} 

    \begin{subfigure}{0.4\textwidth}
        \centering
        \includegraphics[width=\linewidth]{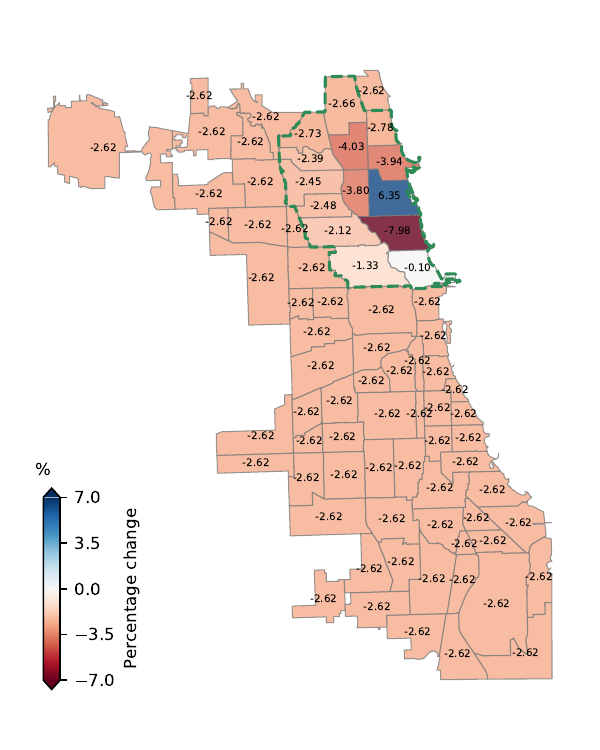}
        \caption{GNN (2 layer).}
        \label{fig:elasticities_gnn2}
    \end{subfigure}
    \hspace{0.0\linewidth} 
    \begin{subfigure}{0.4\textwidth}
        \centering
        \includegraphics[width=\linewidth]{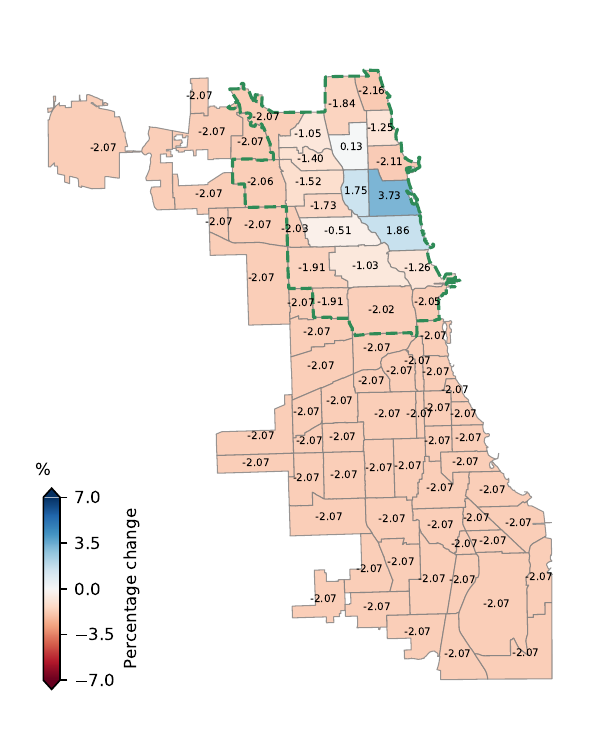}
        \caption{GNN (3 layer).}
        \label{fig:elasticities_gnn3}
    \end{subfigure}

    \caption{Percentage change in the residential choice probability for a selected household following a 10\% increase in housing units in Lake View (Community 6). The boundaries of $k_g$-hop neighbors are indicated by dashed lines.}
    \label{fig:elasticities}
\end{figure}

We further demonstrate the spatial patterns of the cross-elasticities, particularly focusing on how they evolve for GNNs with different layers. To avoid small numbers and for better visualization, Figure.~\ref{fig:elasticities} presents the percentage change w.r.t a 10\% increase in the number of housing units in the Lake View community (community 6), which approximates ten times the elasticities. The boundaries of $k_g$-hop neighbors of GNNs are indicated by dashed lines, which helps to visualize how the influence of neighboring communities varies across different GNN architectures.

As depicted in Figure.~\ref{fig:elasticities}, cross-elasticities of GNNs vary among communities within the $k_g$-hop neighborhood but remain constant for communities outside the $k_g$-hop neighbors, consistent with the theoretical analysis in Section~\ref{sec:substitution}. The MNL model can be viewed as a zero-layer GNN and thus exhibits a constant cross-elasticity across all communities. Additionally, the absolute values of the elasticities appear to decrease as the number of GNN layers increases. This trend arises from the GNN's message-passing mechanism, which introduces a smoothing effect: the change in the utility of a community is ``averaged out'' by its neighbors. Note that the results in Figure.~\ref{fig:elasticities} and Table~\ref{tab:elasticities} are derived from a single household, and specific elasticity values may vary across households. Nevertheless, the observed patterns remain consistent across multiple households. Overall, the elasticity analysis highlights the flexibility of the GNN-DCMs. While the classical MNL is limited by the proportional substitution across the whole map (Figure. ~\ref{fig:elasticities}a), the GNNs can mitigate this constraint by increasing the depth and expanding its reach through the message passing algorithm.

\section{Conclusion and Discussion}
\label{sec:conclusions}
This paper presents a GNN-DCM framework for modeling residential location choice with a large set of spatially correlated alternatives. The GNN-DCM framework provides a deep learning approach for integrating spatial dependencies into discrete choice models. In addition, we demonstrate that the GNN-DCM framework generalizes key classical models, including the NL and SCL models, offering a novel interpretation for DCMs from the perspective of message passing among utilities. The case study of residential location choices in Chicago demonstrates the advantage of the GNN-DCM in predictive performance, design flexibility, capturing individual-level heterogeneity, and spatially-aware substitution patterns.

There are limitations in this paper. This study focuses primarily on developing the GNN-DCM framework and its connections to classical and ANN-based DCMs. As a result, the empirical analysis is intentionally simplified to demonstrate the properties of GNN-DCMs, and thus does not incorporate the full range of household and community-level attributes commonly used in spatial choice modeling (e.g., \citep{sivakumar2007comprehensive, bhat2007comprehensive, hu2019housing}). Additionally, the GNN-DCMs in the case study are designed to enable a fair comparison with baseline models and to highlight the model's spatially-aware substitution patterns. Consequently, the underlying graph is also simplified using only spatial adjacency to construct the adjacency matrix. The full potential of alternative graphs and deep learning techniques, such as using more sophisticated non-adjacency-based graph structure strategies, is not yet fully explored in this study.

Therefore, this work creates new research opportunities by bridging the perspectives in graph, deep learning, and discrete choice models, thus enabling researchers to further synergize classical DCMs with the powerful AI toolbox. From the modeling perspective, future work could investigate other GNN variants and more advanced graph construction techniques, including the use of spatial distance, community similarity, land use patterns, or transportation connectivity to define the graph. There is also potential to explore automatic or learned graph structures, as well as multi-relational graphs that capture multiple types of spatial or functional connections between alternatives. As a general framework, the GNN-DCMs can be widely applied to travel demand modeling tasks, as long as some graph structure exists in alternatives. For example, the GNN-DCM framework could be extended to analyze individual choices among travel modes, land use categories, automobile ownership by incorporating spatial correlation or other network effects \citep{villarraga2025designing, brock2001discrete}. GNN-DCMs can also serve as key modules in more complex joint choice models, such as integrated models of residential location and travel mode choice \citep[e.g.,]{anas1982residential, ben1998integration, bhat2007comprehensive, bhat2010flexible}.

\section*{Acknowledgement}
The authors acknowledges the support from the Research Opportunity Seed Fund 2023 at the University of Florida and the U.S. Department of Energy’s Office of Energy Efficiency and Renewable Energy (EERE) under the Vehicle Technology Program Award Number DE-EE0011186. The views expressed herein do not necessarily represent the views of the U.S. Department of Energy or the United States Government. The authors also acknowledge the early discussions with Dr. Kara Kockelman, Dr. Joan Walker, and Dr. Jinhua Zhao in the research seminars at UT Austin, UC Berkeley, and MIT.

\bibliographystyle{apalike}
\bibliography{reference}

\appendix
\section{Appendix}
\subsection{Proof of proposition~\ref{prop:scl_gnn}}\label{sec:proof_scl_gnn}
\begin{proof}
    The choice probability of the SCL model is:
    \begin{equation*}
        P_{ni} =\frac{\sum_{j \neq i}\left(\alpha_{i, i j} e^{V_{ni}}\right)^{1 / \mu}\left[\left(\alpha_{i, i j} e^{V_{ni}}\right)^{1 / \mu}+\left(\alpha_{j, ij} e^{V_{nj}}\right)^{1 / \mu}\right]^{\mu-1}}
        {\sum_{k=1}^{\left|\V\right|-1} \sum_{l=k+1}^{\left|\V\right|}\left[\left(\alpha_{k, k l} e^{V_{nk}}\right)^{1 / \mu}+\left(\alpha_{l, k l} e^{V_{nl}}\right)^{1 / \mu}\right]^\mu}.
    \end{equation*}
Since $\alpha_{i, i j}=0, \forall j \notin \N(i)$, the choice probability can be rewritten as:
\begin{align}
    P_{ni}&=\frac{\sum_{j \in \N(i)}\left(\alpha_{i, i j} e^{V_{ni}}\right)^{1 / \mu}\left[\left(\alpha_{i, i j} e^{V_{ni}}\right)^{1 / \mu}+\left(\alpha_{j, ij} e^{V_{nj}}\right)^{1 / \mu}\right]^{\mu-1}}
    {\sum_{kl\in \E} \left[\left(\alpha_{k, k l} e^{V_{nk}}\right)^{1 / \mu}+\left(\alpha_{l, k l} e^{V_{nl}}\right)^{1 / \mu}\right]^\mu} \notag\\
    &=\frac{\sum_{j \in \N(i)}\left(\alpha_{i, i j} e^{V_{ni}}\right)^{1 / \mu}\left[\left(\alpha_{i, i j} e^{V_{ni}}\right)^{1 / \mu}+\left(\alpha_{j, ij} e^{V_{nj}}\right)^{1 / \mu}\right]^{\mu-1}}
    {\sum_{kl\in \E} \left[\frac{\left(\alpha_{k, k l} e^{V_{nk}}\right)^{1 / \mu}+\left(\alpha_{l, k l} e^{V_{nl}}\right)^{1 / \mu}}
    {\left(\alpha_{k, k l} e^{V_{nk}}\right)^{1 / \mu}+\left(\alpha_{l, k l} e^{V_{nl}}\right)^{1 / \mu}}\right]
    \left[
    \left(\alpha_{k, k l} e^{V_{nk}}\right)^{1 / \mu}+\left(\alpha_{l, k l} e^{V_{nl}}\right)^{1 / \mu}
        \right]^\mu} \notag\\
    &=\frac{\sum_{j \in \N(i)}\left(\alpha_{i, i j} e^{V_{ni}}\right)^{1 / \mu}\left[\left(\alpha_{i, i j} e^{V_{ni}}\right)^{1 / \mu}+\left(\alpha_{j, ij} e^{V_{nj}}\right)^{1 / \mu}\right]^{\mu-1}}
    {\sum_{kl\in \E}\sum_{ m\in \{k, l\}}
    \left\{\left(\alpha_{m, k l} e^{V_{nl}}\right)^{1 / \mu}
    \left[
    \left(\alpha_{k, k l} e^{V_{nk}}\right)^{1 / \mu}+\left(\alpha_{l, k l} e^{V_{nl}}\right)^{1 / \mu}
        \right]^{\mu-1}
    \right\}}\notag\\
    &=\frac{\sum_{j \in \N(i)}\left(\alpha_{i, i j} e^{V_{ni}}\right)^{1 / \mu}\left[\left(\alpha_{i, i j} e^{V_{ni}}\right)^{1 / \mu}+\left(\alpha_{j, ij} e^{V_{nj}}\right)^{1 / \mu}\right]^{\mu-1}}
    {\sum_{k\in \V}\sum_{ l\in \N(k)}
    \left\{\left(\alpha_{k, k l} e^{V_{nl}}\right)^{1 / \mu}
    \left[
    \left(\alpha_{k, k l} e^{V_{nk}}\right)^{1 / \mu}+\left(\alpha_{l, k l} e^{V_{nl}}\right)^{1 / \mu}
        \right]^{\mu-1}
    \right\}} \notag\\
    &=\frac{\exp \left\{ \log \left( \sum_{j \in \N(i)}\left(\alpha_{i, i j} e^{V_{ni}}\right)^{1 / \mu}\left[\left(\alpha_{i, i j} e^{V_{ni}}\right)^{1 / \mu}+\left(\alpha_{j, ij} e^{V_{nj}}\right)^{1 / \mu}\right]^{\mu-1}\right) \right\} }
    {\sum_{k\in \V}\exp \left\{ \log \left( \sum_{ l\in \N(k)}
    \left(\alpha_{k, k l} e^{V_{nl}}\right)^{1 / \mu}
    \left[
    \left(\alpha_{k, k l} e^{V_{nk}}\right)^{1 / \mu}+\left(\alpha_{l, k l} e^{V_{nl}}\right)^{1 / \mu}
        \right]^{\mu-1} \right)
    \right\}}. \label{eq:from scl to gnn}
\end{align}

Eq.~\eqref{eq:from scl to gnn} shows that the SCL model can be reformulated into the MNL form given by Eq.~\eqref{eq:mnl}, and the exponential terms in both the numerator and the denominator have the same functional form. Continuing to simplify the exponential terms, we have:
\begin{align*}
    &\log \left( \sum_{j \in \N(i)}\left(\alpha_{i, i j} e^{V_{ni}}\right)^{1 / \mu}\left[\left(\alpha_{i, i j} e^{V_{ni}}\right)^{1 / \mu}+\left(\alpha_{j, ij} e^{V_{nj}}\right)^{1 / \mu}\right]^{\mu-1}\right)\\
=&\log \left( \sum_{j \in \N(i)}e^{V_{ni}/\mu + \log(\alpha_{i, i j})/\mu}\left[e^{V_{ni}/\mu + \log(\alpha_{i, i j})/\mu}+e^{V_{nj}/\mu + \log(\alpha_{j, i j})/\mu}\right]^{\mu-1}\right).
\end{align*}
Letting $V^{(0)}_{ni} = V_{ni}/\mu + \log(\alpha_{i, i j})/\mu$, the above expression can be rewritten as a one-layer GNN update function:
\begin{align*}
V_{ni}^{(1)}=&\log \left( \sum_{j \in \N(i)}e^{V^{(0)}_{ni}}\left[e^{V^{(0)}_{ni}}+e^{V^{(0)}_{nj}}\right]^{\mu-1}\right)\\
=&\log \left( \sum_{j \in \N(i)}\exp \left( V^{(0)}_{ni} + (\mu-1)\log\left[e^{V^{(0)}_{ni}}+e^{V^{(0)}_{nj}}\right]\right)\right),
\end{align*}
where $V^{(0)}_{ni}=\frac{V_{ni}}{\mu} + \frac{\log(\alpha_{i, i j})}{\mu} = \frac{\mathbf{b}^{\top}\mathbf{x}_{ni}}{\mu} + \frac{1}{\mu} \log \alpha_{i, ij}$ serves as the initial node representation of the GNN.
\end{proof}

Note that the same proof also holds for generalized spatially correlated logit (GSCL) models \citep{sener2011accommodating}, which takes the same form as the SCL but parameterizes the allocation parameters, $\{\alpha_{i, ij}, \forall j \in \N(i)\}$, by an additional MNL model.

\subsection{Details of GNN update functions}\label{sec:model_details}
The details of the GNN update functions used in the experiments are provided in this section.
\begin{itemize}
    \item \textbf{Message passing neural network (MPNN)}: The update function for MPNN is defined as:
    \begin{equation*}
    \mathbf{h}^{(k)}_{i}=\sigma\left(\mathbf{W} \mathbf{h}_i^{(k-1)} + \bigoplus_{j \in \mathcal{N}(i)} \mathbf{W} \mathbf{h}_j^{(k-1)}\right),
    \end{equation*}
    where $\sigma$ is an element-wise non-linear activation function (we use ReLU \citep{nair2010rectified} for all activations in this study), $\mathbf{W}$ is a learnable weight matrix, and the second term aggregates the representations of neighboring nodes. We test sum, mean, max, and log-sum-exp (LSE) as the aggregation functions $\bigoplus$ in the MPNN update step.
    \item \textbf{Graph convolutional network (GCN)}: The update function for GCN is defined as:
    \begin{equation*}
    \mathbf{h}^{(k)}_{i}=\sigma\left(\sum_{j \in \mathcal{N}(i) \cup \{i\}} \frac{1}{\sqrt{|\mathcal{N}(i)|} \sqrt{|\mathcal{N}(j)|}} \mathbf{W} \mathbf{h}_j^{(k-1)}\right),
    \end{equation*}
    where $\sigma$ is the activation function, $\mathbf{W}$ is a learnable weight matrix, and the normalization term ensures that the updates are invariant to the node degrees.
    \item \textbf{Graph attention network (GAT)}: The update function at each layer computes a weighted sum of a node's neighbor representations, defined as:
    \begin{equation}\label{eq:attention}
    \mathbf{h}^{(k)}_{i}=\sigma\left(\sum_{j \in \mathcal{N}(i) \cup \{i\}} \alpha_{ij} \mathbf{W} \mathbf{h}_j^{(k-1)}\right),
    \end{equation}
    where $\sigma$ is the activation function, $\mathbf{W}$ is a learnable projection matrix, and $\alpha_{ij}$ represents the attention weight indicating the importance of node $j$'s representation to node $i$. The attention weights are calculated by a softmax function over the nodes' representations:
    \begin{equation*}
    \alpha_{ij}=\frac{\exp \left(\text{LeakyReLU} \left(\mathbf{a}^{\top}\left[\mathbf{W} \mathbf{h}_i || \mathbf{W} \mathbf{h}_j\right]\right)\right)}{\sum_{k \in \mathcal{N}(i)\cup \{i\}} \exp \left(\text{LeakyReLU}\left(\mathbf{a}^{\top}\left[\mathbf{W} \mathbf{h}_i || \mathbf{W} \mathbf{h}_{k}\right]\right)\right)},
    \end{equation*}
    where $||$ denotes concatenation, and $\mathbf{a}$ is a learnable coefficient vector. It is often to use separate Eq.~\eqref{eq:attention} multiple times (i.e., stacking the outputs of multiple functions), referring to the multi-head attentions. We choose the number of heads to be 1 when the hidden dimension $h=1$ and 8 otherwise.
\end{itemize}

\subsection{Details of skip connections}\label{sec:skip_connections}
Skip connection was first popularized by the ResNet architecture \citep{he2016deep}. It connects the input of a layer to the output of a deeper layer, enabling the training of very deep neural networks by mitigating the vanishing gradient problem. In GNNs, skip connections are implemented in various ways, such as through summation \citep{he2016deep} or concatenation \citep{huang2017densely}. These approaches serve the same purpose—preserving the original node representation—and typically have a similar performance in GNNs \citep{you2020design}.

In this paper, we adopt the gated skip connection via summation \citep{srivastava2015highway}. For example, the update function of a GNN with skip connections for MPNN is:
\begin{equation*}
\mathbf{h}^{(k)}_{i}=\sigma\left((1-c) \mathbf{h}_i^{(k-1)} + c\left(\mathbf{W} \mathbf{h}_i^{(k-1)} + \bigoplus_{j \in \mathcal{N}(i)} \mathbf{W} \mathbf{h}_j^{(k-1)}\right) \right),
\end{equation*}
where $c=\mathrm{sigmoid}(\mathbf{W}_c \mathbf{h}_i^{(k-1)} + \mathbf{b}_c)\in (0,1)$ is a learnable gate parameter, which allows the model to adaptively balance between retaining the original node representation and incorporating the output from the GNN message passing. This layer reverts to which of the ASU-DNN when $c\rightarrow0^+$. In addition, it keeps the same dimensionality across hidden layers. We use the same skip connection mechanism for all GNNs used in this paper, including the MPNN, GCN, and GAT.

\end{document}